\documentclass[journal]{IEEEtran}
\ifCLASSINFOpdf
\else
\fi
\usepackage[utf8]{inputenc} % allow utf-8 input
\usepackage[T1]{fontenc}    % use 8-bit T1 fonts
\usepackage{hyperref}       % hyperlinks
\usepackage{url}            % simple URL typesetting
\usepackage{booktabs}       % professional-quality tables
\usepackage{amsfonts}       % blackboard math symbols
\usepackage{nicefrac}       % compact symbols for 1/2, etc.
\usepackage{microtype}      % microtypography

\usepackage{soul}
\usepackage{color}
\usepackage{algorithm}      % algorithm
\usepackage[noend]{algpseudocode} % algorithm
\usepackage{amsmath}

\usepackage{graphicx}
\usepackage{subcaption}

\usepackage[export]{adjustbox}
\usepackage{lineno}
\usepackage{enumitem}

\usepackage{setspace}

\usepackage{todonotes}
\usepackage{lipsum}% http://ctan.org/pkg/lipsum

\usepackage{balance}

\usepackage{hyperref}       % hyperlinks
\usepackage{url}            % simple URL typesetting

\usepackage{graphicx,subcaption}
\usepackage{color}
\usepackage{soul}
\usepackage{amsmath}
\usepackage{cleveref}
\usepackage{algorithm,algpseudocode}
\usepackage{mathptmx} % assumes new font selection scheme installed
\usepackage{multirow}
\usepackage{array}
\usepackage{booktabs}
\usepackage{pifont}
\usepackage{balance} 
\usepackage{blindtext}

\usepackage{tikz}

\usepackage{setspace}
\usepackage{amsfonts}
\usepackage{amsmath}
\usepackage{newtxtext, newtxmath}

\begin{document}
\title{Towards Cognitive Navigation: Design and Implementation of a Biologically Inspired Head Direction Cell Network}

\author{Zhenshan Bing$^1$, Amir EI Sewisy$^{1}$, Genghang Zhuang$^1$, Florian Walter$^1$, \\Fabrice O. Morin$^{1}$, Kai Huang$^{2,*}$ and Alois Knoll$^1$
	\thanks{Authors' Affiliation: $^1$ Department of Informatics, Technical University of Munich, Germany. $^2$ Key Laboratory of Machine Intelligence and Advanced Computing (Sun Yat-sen University), Ministry of Education, China; School of Data and Computer Science, Sun Yat-sen University, China. }
	%\thanks{Address: \dag Boltzmannstraße 3, 85748 Garching bei M\"unchen (Germany).}
	\thanks{Email: $\{$bing,~zhuang,~walter,~morinf,~knoll$\}$@in.tum.de, amir.el-sewisy@tum.de}%
	\thanks{*Corresponding author: huangk36@mail.sysu.edu.cn}
	%	\thanks{Email: bing@in.tum.de,~jiangz@in.tum.de,~chenglong3@mail.sysu.edu.cn,~huangk36@mail.sysu.edu.cn, knoll@in.tum.de}%
}


\markboth{Journal of \LaTeX\ Class Files,~Vol.~14, No.~8, August~2015}%
{Shell \MakeLowercase{\textit{et al.}}: Bare Demo of IEEEtran.cls for IEEE Journals}



\maketitle

\begin{abstract}
As a vital cognitive function of animals, the navigation skill is first built on the accurate perception of the directional heading in the environment.
Head direction cells (HDCs), found in the limbic system of animals, are proven to play an important role in identifying the directional heading allocentrically in the horizontal plane, independent of the animal's location and the ambient conditions of the environment.  
However, practical HDC models that can be implemented in robotic applications are rarely investigated, especially those that are biologically plausible and yet applicable to the real world.  
In this paper, we propose a computational HDC network which is consistent with several neurophysiological findings concerning biological HDCs, and then implement it in robotic navigation tasks.
The HDC network keeps a representation of the directional heading only relying on the angular velocity as an input. 
We examine the proposed HDC model in extensive simulations and real-world experiments and demonstrate its excellent performance in terms of accuracy and real-time capability.
\end{abstract}


% \fi
\IEEEpeerreviewmaketitle

\section{Introduction}
\label{sec:intro}

\IEEEPARstart{T}{he} ability to localize oneself and update one's position based on self-motion cues in a spatial environment is one of the most fundamental cognitive functions of animals.
Relying on their spatial cognitive navigation skills, animals can successfully forage, migrate, and mate.
Many animals are naturally born with powerful and yet efficient navigation skills, such as the echolocation by bats, magnetic-based location~\cite{riley1957echolocation} by migratory birds~\cite{mouritsen2004cryptochromes}, and the most common vision-based navigation.
These various skills usually outperform state-of-the-art artificial technologies.  
Studies from neuroscience have revealed that the hippocampus in the brain plays an important role in spatial navigation by coordinating several types of neurons with different functionalities~\cite{o1978hippocampal}, such as head direction cells (HDCs)~\cite{ranck1984head}, grid cells (GCs)~\cite{derdikman2009fragmentation}, and place cells (PCs)~\cite{diba2007forward}.

As the foundation of a successful navigation behavior, HDCs are found to fire in relation to the animal's directional heading with respect to the environment.
The neural activity of a given HDC will be much higher when the agent is facing at a constant preferred direction over time in the entire environment.
In another word, the HDC network resembles the functions of a compass without relying on the geomagnetic field of the earth.
Over the past few decades, studies on HDCs in the fields of neuroscience and computer science have designed several biologically realistic HDC models~\cite{taube1990head,zhang1996representation,arleo2000modeling,taubeHeadDirectionSignal2007} and implemented them on robotic systems~\cite{zhou2017robot, kreiserNeuromorphicApproachPath2018,zengNeuroBayesSLAMNeurobiologicallyInspired2020}.
These models were usually built on the basis of one-dimensional continuous attractor network to integrate the angular velocity of the robot and form a stable representation of the directional heading~\cite{wu2016continuous}.

However, most HDC models focus mainly on biological
model fidelity which makes them computationally expensive
and inaccurate in robotic applications.
For instance, Zhou et al. proposed an approach to generate place cells and head-direction cells using an unsupervised learning algorithm based on slow feature analysis \cite{zhou2017robot}. 
The averaged error of their HDCs over all directions was around $14.73^{\circ}$.
%Arleo et al. proposed a cognitive navigation system which consisted of coupled HDCs and PCs~\cite{arleo2000modeling}.  
%In their work, the uncalibrated mean tracking error of the HDC controller quickly went up to $60^{\circ}$ after turning 150 time steps.
Hence, the design of a HDC model with biological plausibility and applicability is challenging.
The reasons are multifold. 
First, the mechanisms of HDCs are not fully understood by researchers and thus a proper model to imitate the dynamics of HDCs is missing. 
\textcolor{black}{
	For example, some HDC models take spikes at the cost of high computational complexity.}
As a consequence, they can not be applied in real-time tasks. 
Second, the shifting mechanism of the represented directional heading \textcolor{black}{that is used to modulate the activity profile of} the HDC network is not clear.
Third, the HDC network needs time to compute the operating patterns to match with \textcolor{black}{the high sampling frequency of the input sensory data}.
For field robotic navigation tasks, this computation is usually performed on MCUs (micro control units) with limited computing power, and the length of the computing time will directly affect the accuracy of representing the directional heading.

This work provides a HDC based network and that keeps a relatively accurate representation of the directional heading only relying on the angular velocity as an input. 
The proposed HDC network is proven to be accurate and applicable to be used in real-world robotic implementations. 
The contributions of this work are summarized as follows.
\begin{itemize}
	\item We propose a discrete formulation of the dynamics of HDCs on the basis of the latest neuron model that is supported by data recorded from in-vivo neurons and yet being lightweight. 
	This discrete HDC representation can enable accurate and fast computation of the directional heading in real-world implementation.
	
	\item We propose a continuous dynamic shifting mechanism of the activity profile of the HDC network. 
	This shifting mechanism can ensure smooth and accurate activity transitions without disturbing the shape of the profile.
	
	\item To demonstrate the accuracy and robustness of the proposed HDC model, we run extensive experiments including navigating a robot in both simulated and real-world scenarios as well as testing using the widely used dataset KITTI~\cite{geiger2013vision}.
	Experiment results showed higher accuracy of estimating the directional heading of the robot than the previous work and better robustness than the method that directly integrates the angular velocity.
	The online testing experiment also proved the real-time capability of the proposed HDC network even on a Raspberry Pi 3.
	
\end{itemize}

\section{Background}

\subsection{Biological Background}
In 1979, John O'Keefe and Lynn Nadel discovered that rats could store cognitive maps in the hippocampus~\cite{o1979precis}, which was vital for processing spatial information with the help of multiple classes of neurons with different functionalities.
Head directional cells were found in 1983~\cite{ranck1985head} and named for the fact that each cell responds to different head directions with different firing rates when the head of a freely moving rat points in a restricted range of angles in the horizontal plane.
HDCs are typically remarkably insensitive to head movement~\cite{blair1995anticipatory}. 
Such cells were also recorded in multiple different species, for example, rats~\cite{ranck1985head} and drosophilia (fruit flies)~\cite{kim2017ring}.
\textcolor{black}{The activities of HDCs are mainly influenced by one's directional heading and can be calibrated by external cues, for instance, using visual landmarks.
} 
It was also shown that HDCs remain functional in the dark \cite{taube2007head}, which means the cell's functionality can be maintained without visual cues, only using self-movement information.

\subsection{General Properties}
An idealized HDC's firing rate can be defined as a function of the current head direction, and this function or mapping is referred as the HDC's tuning curve~\cite{taube1990head}.
Each HDC can be characterized by different parameters of its turning curve. 
The peak of the tuning curve, i.e., the head direction associated with the highest activity (firing rate), is referred to as the cell's preferred direction~\cite{taube1990head}, which can range from $0^{\circ}$ to $360^{\circ}$.
The activity of each HDC plotted over their preferred directions will be referred to as the HDC’s activity profile.
The activity profile of a biological HDC roughly resembles the shape of a Gaussian bell curve~\cite{taube2007head}. 
Each HDC's firing rate is mostly maintained at a very low level when its preferred direction is in a direction away from the current directional heading of the animal, and then increases dramatically to its maximum firing rate when the animal moves its head very close to the preferred direction of the HDC.

\subsection{Continuous Attractor Networks}
Over the past decades, the continuous attractor networks (CANs) have been widely used to represent the HDC networks, which enable persisting and tracking the directional heading in a circle over time.
The CAN model consists of a value space which is a cell population representing the directional heading and facilitates a winner-take-all (WTA) mechanism, referred to as the attractor topology which ensures that at any time only one peak of neural activity can be activated.
One important characteristic of CANs are their translation-invariant connections between neurons, i.e., the connection strength between two neurons is only determined by the difference between their respective represented states in the value space~\cite{wu2016continuous}.
Normally, there will be a shifting mechanism that enables shifts of the peak activity in the value space in a direction prescribed by the corresponding stimulus applied to the network.
Several kinds of CANs have been proposed in the literature and their dynamics were mathematically analyzed~\cite{brody2003basic, wu2008dynamics}.
We also build up our HDC network on the basis of the concept of continuous attractor networks.

\section{Related Work}

There are many studies that applied biologically inspired HDC models in real-world robotic tasks, in which the accuracy of the estimated directional headings of those agents played decisive roles on their final performance of the tasks.

Arleo et al. proposed a biological navigation system including a strongly coupled HDC network and PC network.
Their HDC network modeled three neural populations, namely, the head angular velocity cells, the lateral mammillary nuclei direction cells, and the anterodorsal thalamic nucleus cells~\cite{arleo2000modeling}. 
The uncalibrated mean tracking error of their HDC model quickly went up to $60^{\circ}$ after turning $150$ time steps.
Even integrated with a visual-based calibration mechanism, the error was still around $10^{\circ}$.
Zhou et al. proposed an approach to generate place and head-direction cells using an unsupervised learning algorithm based on slow feature analysis \cite{zhou2017robot}. 
The averaged error of their HDCs over all directions was around $14.73^{\circ}$.
Degris et al. developed a HDC model with similar architecture as \cite{arleo2000modeling}, but based on spiking neurons \cite{degrisSpikingNeuronModel2004}.
Their spike-based HDC model was implemented on a mobile car and showed an averaged error of $9^{\circ}-12^{\circ}$.

There are also a number of studies that used biologically inspired HDCs as part of their navigation controller together with other functional neurons, such as place cells and grid cells.
One of the most widely investigated topics is RatSLAM~\cite{milfordRatSLAMHippocampalModel2004a, milfordMappingSuburbSingle2008a} and its related correlational studies~\cite{zengNeuroBayesSLAMNeurobiologicallyInspired2020, tangCognitiveNavigationNeuroInspired2018}.
RatSLAM \cite{milfordRatSLAMHippocampalModel2004a} is a Simultaneous Localization and Mapping (SLAM) solution based on neuronal mechanisms in the rat brain and tries to replicate the phenomenon of place fields of rat neurons.
RatSLAM used competitive attractor networks to keep stable activity packets to store the belief about position and orientation, i.e., the Pose Cell network. 
Their pose cells were similar to a combination of the biologically observed HDCs and GCs.
The focus in RatSLAM lied on compensating odometry errors using visual cues.
Unlike the HDC network model presented in this work, the activity in RatSLAM’s Pose Cells was not shifted by neuronal mechanisms but simply by  directly moving the activity packet.

\section{Models and Network Architecture}
\label{sec_model}

In this section, we will first describe the architecture of our HDC network and then present the model of the HDC neurons, synapses, and the decoding strategy of the HDC network.

\begin{figure}[t]
	\centering
	\includegraphics[width=0.40\textwidth, trim={0.5cm 5cm 0.8cm 2cm},clip]{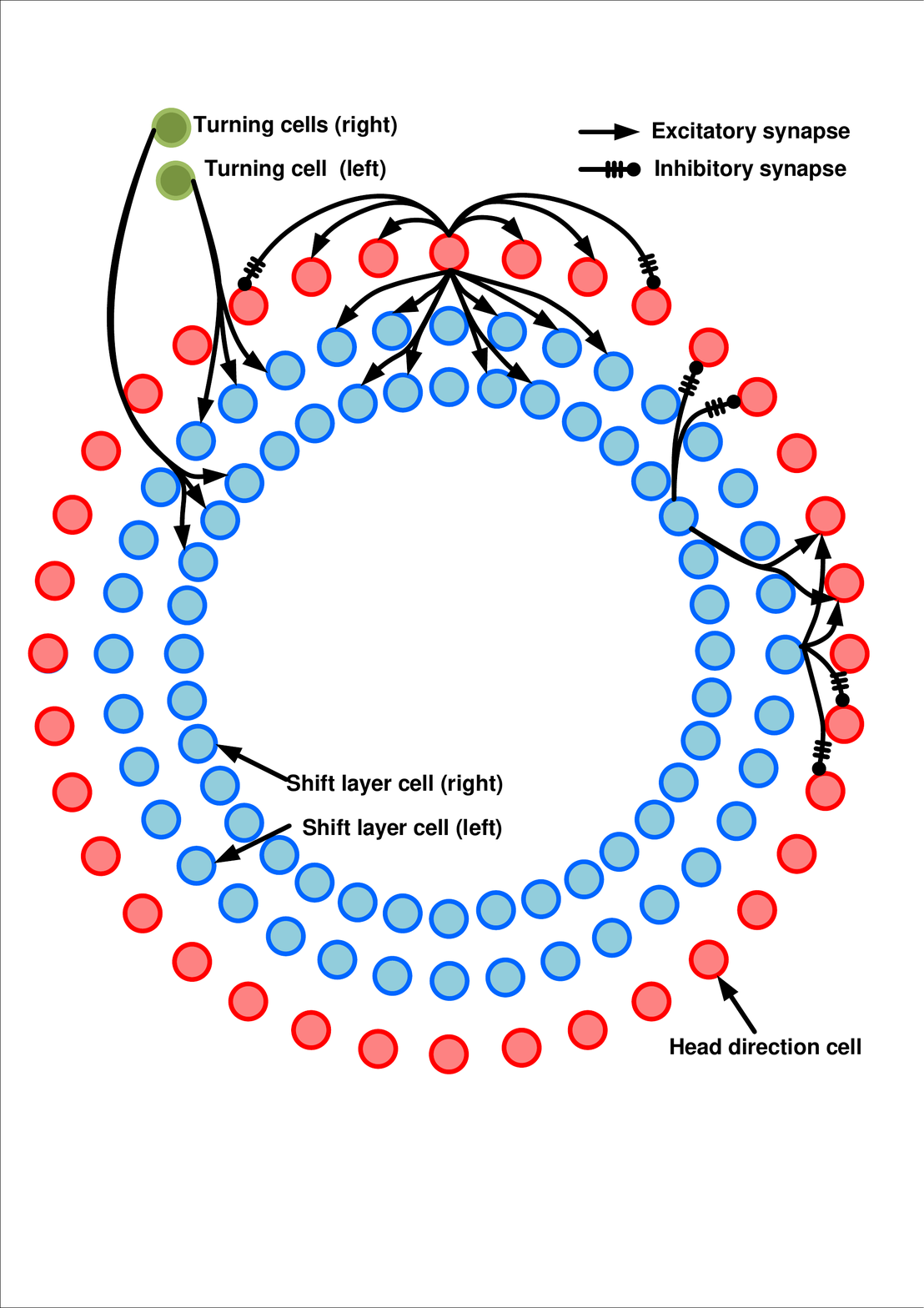}
	\caption{Schematic illustration of the HDC network inspired by \cite{skaggsModelNeuralBasis1995}. 
		\textcolor{black}{There are three different layers of neurons, namely, the HDC layer, the turning left layer, and the turning right layer. Each layer contains $100$ neurons.}
		For simplicity, only a part of the outgoing synaptic connections of one HDC neuron, two shifting neuron, and turning neurons are shown.}
	\label{fig:topology}
\end{figure}

\subsection{Attractor Topology}

\begin{figure}[!tb]
	\centering
	\includegraphics[width=0.48\textwidth]{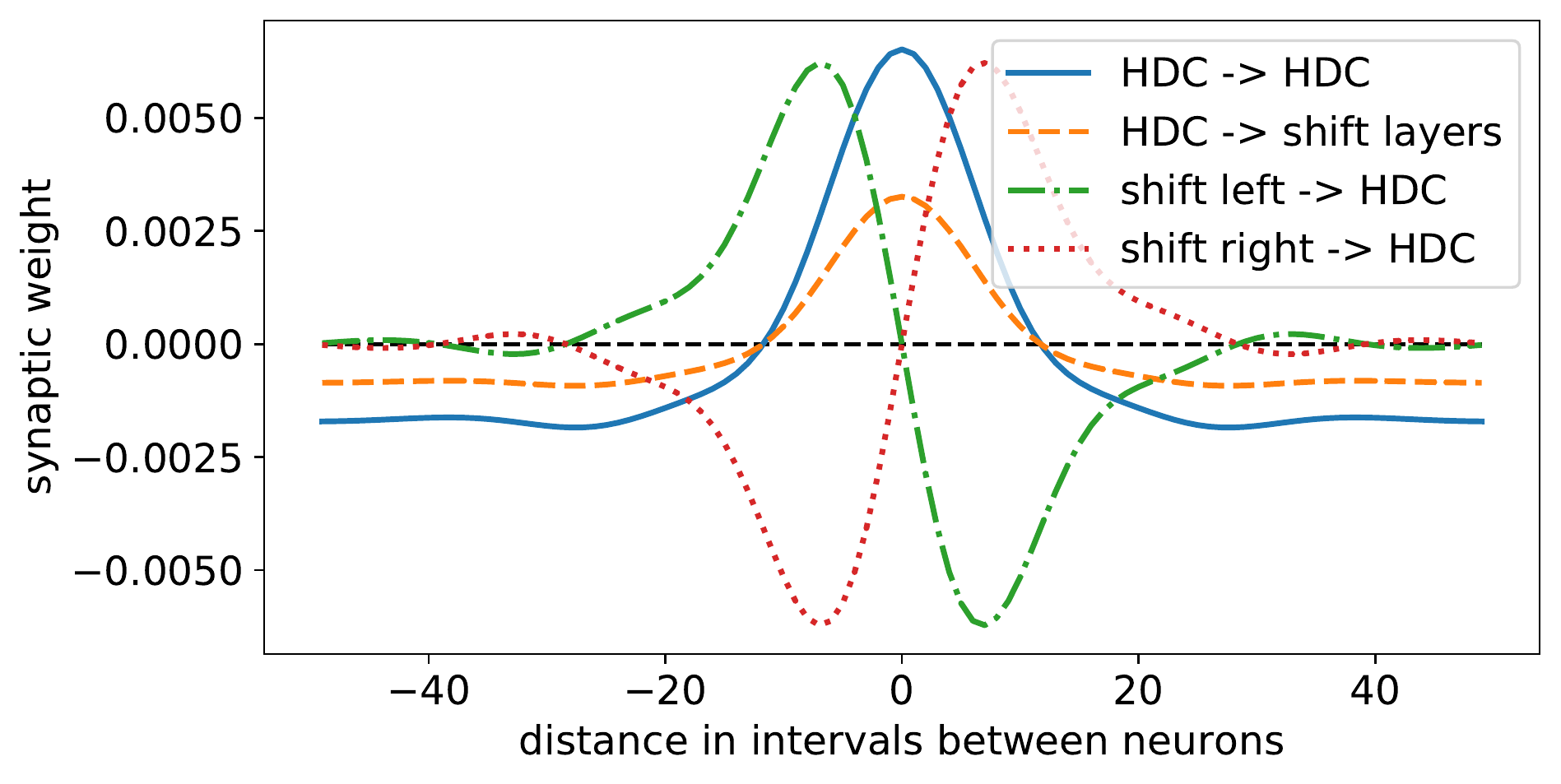}
	\caption{All synaptic weights in the HDC network. The synaptic weights are defined as functions of the distance given in intervals between cells. The positive distance corresponds to right (clockwise) neighbors and the negative distance corresponds to left (counterclockwise) neighbors.
		Positive weights mean excitatory synapses while negative weights means inhibitory synapses. }
	\label{fig:all_weights}
\end{figure}

On the basis of CANs, our HDC network is built up and visualized in Figure~\ref{fig:topology}.
The HDC network consists of three layers, namely, the HDC layer (red solid circles) and two shift layers (blue solid circles) to shift the peak activity of the HDC layer to the left or right direction.
Additionally, two turning cells are used to inject turning stimuli to the HDC network.
The HDC layer represents the directional heading and consists of $n=100$ head direction cells.
Their preferred direction $\theta_i$ ($i \in [0, ..., n-1]$) are equally  distributed around the circle and are given in radians in the interval $[0,~2\pi]$.
The HDC $i$ has the preferred direction $\theta_i = \frac{2\pi i}{n}$.
The connections within the HDC layer are set up so that neighbouring cells are connected with excitatory synapses while distant cells are wired with inhibitory synapses.
\textcolor{black}{Therefore, for one HDC, it has $99$ synapse connections from the other HDCs and $100$ synapse connections from each shift layer in this paper.}
\textcolor{black}{An intuitive illustration of the excitatory and inhibitory connections is shown in Figure \ref{fig:all_weights}, where positive synaptic weight means excitation and negative synaptic weight means inhibition.}
Each cell in the shift layer is associated with one cell in the HDC layer and its activity closely follows this HDC but with a lower peak firing rate. 
Take one shift left cell as an example, it is connected to the HDCs on the left side of its corresponding HDC with excitatory synapses and to the other half with inhibitory synapses. 
There is no stimulus input from the shift left cell to its corresponding HDC.
The turning left cell injects stimuli to all the cells in the shift left layer and the same rule is applied to the turning right cell. 
On the one hand, the overall directional heading in the HDC layer is calculated by averaging all the activities with the \textit{population vector coding} rule \cite{georgopoulos1986neuronal}.
On the other hand, a shifting mechanism enables shifts of the peak activity in the HDC layer in a direction prescribed by the corresponding stimulus applied to the network.

\begin{figure}[!tb]
	\centering
	\includegraphics[width=0.45\textwidth]{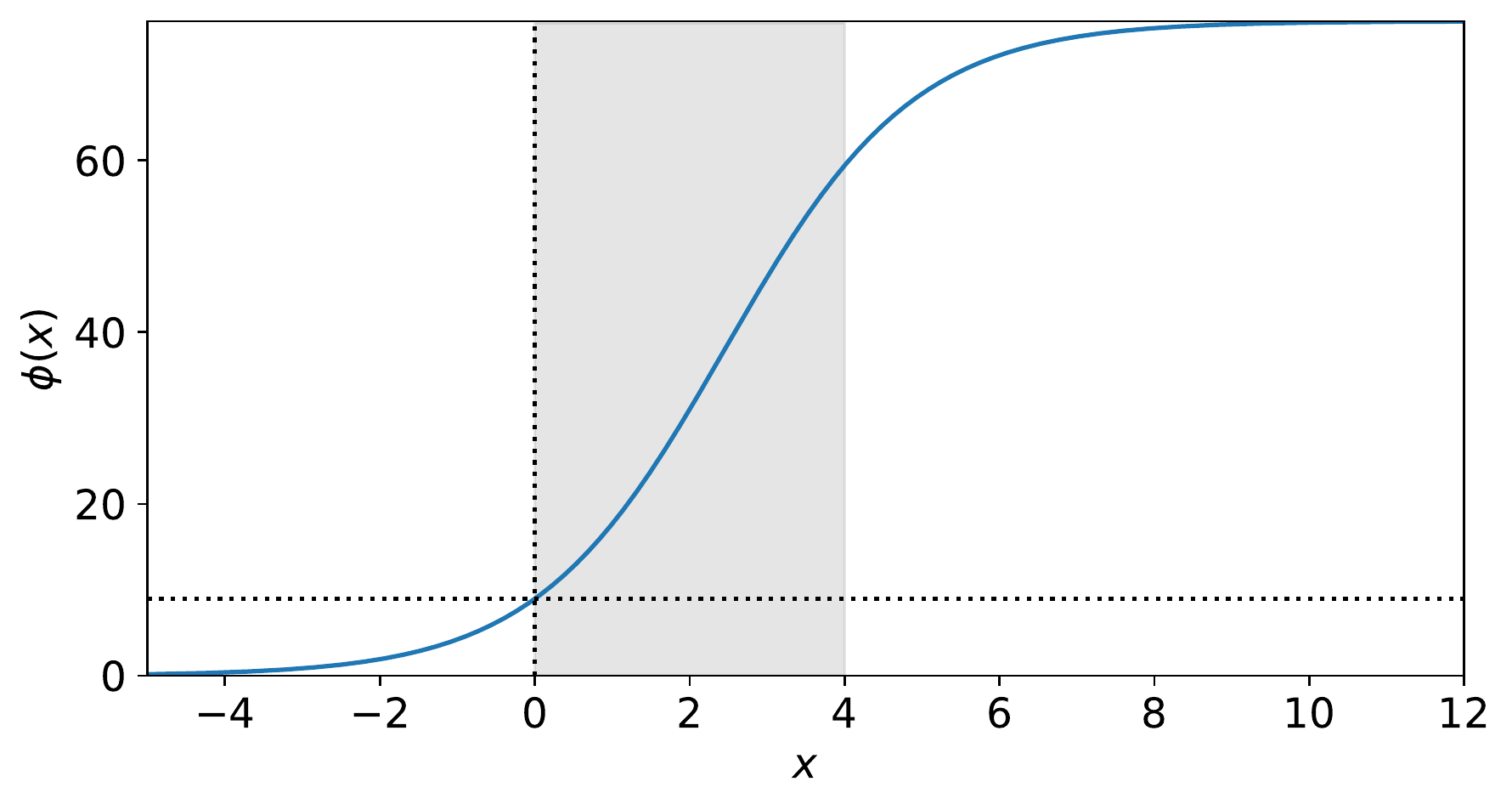}
	\caption{The single-neuron transfer function $\phi(x)$.}
	\label{fig:phi}
\end{figure}

\subsection{HDC Model}
\label{sub_HDC_Neuron}
Biological neurons fire spikes which influence other neurons connected by synapses. 
\textcolor{black}{
	Here, a neuron model inspired by \cite{PEREIRA2018227} is used.} 
This model is selected because it is modeled using data recently recorded from biological neurons.
The firing rate $f$ is governed by the standard rate equation:
\begin{equation}
	\tau_1\frac{df_i}{dt}=-f_i+\phi(I_i+\sum_{j=1...n,~i\neq j}w_{ij}f_j)\text{,}
	\label{eq:time_evolution_neuron}
\end{equation}
where $\tau_1 = 20$ ms is the time constant of the firing rate dynamics, $\phi$ is the transfer function from the synaptic current to the firing rate.
$w_{ij}$ is the synaptic strength, also referred to as synaptic weight, from neuron $j$ to neuron $i$. 
Excitatory synapses are modeled with positive weights and inhibitory synapses are represented with negative weights. $I_i$ is the external input current to neuron $i$.
%According to \eqref{eq:time_evolution_neuron}, the stable equilibrium of firing rate $f$ will converge to $\phi(\cdot)$.

The transfer function $\phi$ is modeled as a sigmoid function:
\begin{equation}
	\phi(x)=\frac{r_m}{1+e^{(-\beta_t(x-h_0))}}
	\label{eq:phi_neuron}
\end{equation}
where $r_m = 76.2$ Hz is the maximal firing rate. 
$\beta_t = 0.82$ and $h_0 = 2.46$ are the shape parameters to adjust the slope and shift of $\phi$.
The parameters $r_m$, $\beta_t$ and $h_0$ were inferred to fit data recorded from in-vivo neurons by \cite{PEREIRA2018227}.
$\phi(x)$ is visualized in Figure \ref{fig:phi}.
Following \eqref{eq:time_evolution_neuron}, the firing rate of a neuron with constant external and synaptic inputs $x$ converges to $\phi(x)$ over time. 
Note that $\phi(0)\approx8.95$, which means that an isolated neuron fires at a rate of about 8.95 Hz without any external input. 

%
%\subsection{Tuning Curve}
The mapping of the head direction $\theta$ to a HDC's activity (firing rate $f$) is referred as its tuning curve.
The typical direction tuning curve of each HDC is usually modeled as a Gaussian-like shape function, which is fitted as 
\begin{equation}
	f(\theta) = A + Be^{M \cos(\theta - \theta_0)}\text{,}
	\label{eq_3}
\end{equation}
where $A$, $B$, and $M$ are constant parameters. 
$\theta_0$ is the preferred direction of each HDC.
Combining the properties of the HDC model described in \eqref{eq:time_evolution_neuron} and \eqref{eq:phi_neuron}, which fires at $8.95$ Hz without any external stimulus and at 76.2 Hz as the maximal firing rate, we can select the parameters for the turning curve, where $A = 8.95$, $B \approx0.344$, and $M \approx 5.29$.
\textcolor{black}{It should be noted that the proposed HDC model is also used to construct neurons in both shift layers.}

\subsection{Synapse Model}
\textcolor{black}{
	We introduce the synapse model and how to calculate the synaptic weights in this section.}
For the HDC network with the same preferred direction $\theta$, we use $f(\theta, t)$ to represent the averaged firing rate of all HDCs with the preferred direction $\theta$ at time $t$.
$w(\Delta\theta, t)$ is referred to as the weight distribution function and is defined as the synaptic weight between HDCs with their preferred directions differing by $\Delta\theta$.
Then according to \cite{Hopfield3088}, there is a resistance-capacitance charging equation that determines the rate of change of the total synaptic input $u(\theta, t)$ as 
\begin{equation}
	\begin{split}
		\tau_2\frac{du_i(\theta, t)}{dt} = -u_i(\theta, t)+w(\Delta\theta, t)*f(\theta, t)
	\end{split}
	\label{eq:time_evolution_zhang}
\end{equation}
\textcolor{black}{$\mu$ is used to represent the total synaptic input for the neuron with preferred direction $\theta$ and the synaptic input from one single neuron can be calculated as $w \times f$, where $w$ is the synaptic weight and $f$ is the firing rate of the neighbouring neuron defined in \eqref{eq_3}.}
$\tau_2=10$ms is a constant and $w*f$ denotes the convolution of $w(\Delta\theta, t)$ and $f(\theta, t)$. 
This convolution is utilized to to make the system rotation-invariant, which means the connectivity between two cells with different preferred directions $\theta_1$ and $\theta_2$ depends only on their distance $\Delta \theta = \theta_1-\theta_2$. 
The convolution is defined by~\cite{zhang1996representation} as:
\begin{equation}
	w(\Delta \theta, t)*f(\theta, t)=\frac{1}{2\pi}\int_{0}^{2\pi}w(\Delta \theta-\alpha, t)f(\alpha, t)d\alpha
	\label{eq:convolution_zhang}
\end{equation}
According to~\cite{zhang1996representation}, the weights yielding stable states close to a given firing rate profile $f$ can be calculated as 
\begin{equation}
	\hat{w}_n=\frac{\hat{u}_n\hat{f}_n}{\lambda+\hat{f}_n^2}
	\label{eq:inferweights}
\end{equation}
$\hat{f}_n$, $\hat{w}_n$, and $\hat{u}_n$ correspond to the $n^{th}$ Fourier coefficient of $f$, $w$, and $u$, respectively. 
A brief mathematical derivation is given in Appendix~\ref{appendix_1}.
The parameter $\lambda$ controls the flatness of the solution $w$.
By minimizing the error between the target activity profile and the simulated one, we choose $\lambda=25824$.

\begin{algorithm}[!b]
	\caption{Weights calculation}
	\label{algo_w}
	\begin{algorithmic}[1]
		\renewcommand{\algorithmicrequire}{\textbf{Input:}}
		\renewcommand{\algorithmicensure}{\textbf{Output:}}
		\Require \text{}
		
		$n$: number of neurons 
		
		$\theta_i$: preferred directions
		
		$\lambda$: flatness parameter
		
		$A, B, K$: parameters for target firing rates
		
		\textcolor{black}{$\phi$: transfer function defined in \eqref{eq:phi_neuron}}
		
		\textcolor{black}{$\mathcal{F}$ and $\mathcal{F}^{-1}$ are the Fast Fourier Transform solver and inverse Fast Fourier Transform solver \cite{virtanen2020scipy}}
		
		\Ensure \text{}
		
		$W$: weights indexed by the distance in HDCs 
		
		\For{$i=0...n-1$}
		\State $F[i]:=A+Be^{Kcos(\theta_i)}$
		\Comment{\textcolor{black}{calculate target firing rates $F$ as \eqref{eq_3}}}
		\EndFor
		\For{$i=0...n-1$}
		\State $U[i]:=\phi^{-1}(F[i])$
		\Comment{\textcolor{black}{calculate $U$ from $F$}}
		\EndFor
		\State $\hat{F}:=\mathcal{F} (F)$; 
		\Comment{\textcolor{black}{apply Fast Fourier Transform}}
		\State $\hat{U}:=\mathcal{F} (U)$;
		\Comment{\textcolor{black}{apply Fast Fourier Transform}}
		\For{$i=0...n-1$}
		\State $\hat{W}[i]= \frac{\hat{U}[i]\hat{F}[i]}{\lambda+ |\hat{F}[i]|^2}$
		\Comment{calculate Fourier coefficients as \eqref{eq:inferweights}}
		\EndFor
		\State $W:=\mathcal{F}^{-1} (\hat{W})$
		\Comment{\textcolor{black}{apply inverse Fast Fourier Transform}}
		\State return W;
	\end{algorithmic}
\end{algorithm}

\textcolor{black}{To calculate the synaptic weight $\hat{w}_n$, we also need to calculate the total synaptic input $\hat{u}_n$.}
As described in \eqref{eq:time_evolution_neuron}, the desired firing rate profile $f$ will finally converge to $\phi(\sum_{j}w_{ij}f_j)=\phi(u_i)$ over time.
Then we can get 
\begin{equation}
	u_i:=\phi^{-1}(f_i)
	\label{eq_cal_u}
\end{equation}
For discrete HDCs, the synaptic weight function $w'(\Delta \theta)$ can be defined as $w_{ij}=:w'(\theta_{ij})$, where $\theta_{ij}$ is the distance between the preferred directions of cell $i$ and cell $j$. 
Thus, $w'$ will be used instead of $w$ in \eqref{eq:inferweights}. 
\textcolor{black}{$\hat{u}_n$ is calculated as \eqref{eq_cal_u}.
	Then, the discrete version of \eqref{eq:inferweights} is as follows}
\begin{equation}
	\hat{w'}_n=\frac{\hat{u}_n\hat{f}_n}{\lambda+|\hat{f}_n|^2}
	\label{eq:inferweights_adapted}
\end{equation} 
\textcolor{black}{First, $\hat{f}_n$ is calculated by getting the $n^{th}$ Fourier coefficient of the vector of all target firing rates $F:=(f_i)_i$.
	$\hat{u}_n$ is calculated by getting the $n^{th}$ Fourier coefficient of the vector $U:=(\phi^{-1}(f_i))_i$.} The obtained $\hat{w}_{n}$ is interpreted as the Fourier coefficient of the vector $W$. 
The $i^{th}$ element of $W$ is the weight between HDCs which are $i$ steps apart. 
With a zero-based numbering vector, the $0^{th}$ element of $W$ is the weight from a cell to itself, the $1^{st}$ element of $W$ is the weight between direct neighbors, and so forth.
The pseudocode for calculating $W$ is given in Algorithm~\ref{algo_w}.
The synaptic weights connecting the cells in the HDC layer are visualized by the solid curve in Figure~\ref{fig:all_weights}.
In the HDC layer, there are strong excitatory connections between neighboring cells, and strong inhibitory connections between distant cells.

\subsection{Angle Decoding}
To interpret the movement of the HDC network's activity peak, the directional heading needs to be decoded from the HDC network. 
We apply the \textit{population vector coding}~\cite{georgopoulos1986neuronal}, by taking the average angle of the preferred direction $\theta_i$ weighted by the respective firing rates $f_i$ for every HDC. 
The decoded directional heading $\bar{\theta}$ is simply obtained as
\begin{equation}
	\bar{\theta} = arctan \big(\frac{\sum^n_{i=0} sin(\theta_i) f_i(t) }{\sum^n_{i=0} cos(\theta_i) f_i(t)} \big)
	\label{eq:decodeDirection}
\end{equation}
%
%
%\subsection{Attractor Network Initialization}

%\begin{algorithm}[t]
%	\caption{Algorithm for ...}
%	\begin{algorithmic}[1]
%%		\renewcommand{\algorithmicrequire}{\textbf{Input:}}
%%		\renewcommand{\algorithmicensure}{\textbf{Output:}}
%		\REQUIRE \text{}
%		\\ $n$: number of neurons \\
%		$\theta_i$: preferred directions\\
%		$\lambda$: flatness parameter\\
%		$A, B, K$: parameters for target firing rates
%		\ENSURE  out
%		\\  $W$: a list of weights indexed by the distance in HDCs \\
%		\textit{\# calculate target firing rates $F$}\\
%		\FOR {$i=0...n-1$}
%		\STATE $F[i]:=A+Be^{Kcos(\theta_i)}$
%		\ENDFOR \\
%		\textit{\# calculate U from F}\\
%		\FOR{$i=0...n-1$} 
%		\STATE	$U[i]:=\phi^{-1}(F[i])$
%		\ENDFOR \\
%		\textit{\# calculate fourier transforms}\\
%		$\hat{F}:=fft(F)$\\
%		$\hat{U}:=fft(U)$\\
%		\textit{\# apply Equation \ref{eq:inferweights_adapted} to get the fourier coefficients of W}\\
%		
%		\FOR{$i=0...n-1$} 
%		\STATE	$\hat{W}[i]= \frac{\hat{U}[i]\hat{F}[i]}{\lambda|\hat{F}[i]|^2}$
%		\ENDFOR \\
%		\textit{\# apply the inverse fourier transformation to $\hat{W}$ to get W}\\
%		\STATE $W:=ifft(\hat{W})$ \COMMENT {test}
%		\STATE return W
%		\caption{Calculate weights with $\lambda$ given}
%		\label{alg:weightfunction}
%	\end{algorithmic} 
%\end{algorithm}

\begin{figure}[t]
	\centering
	\includegraphics[width=0.48\textwidth, trim={5cm 7.5cm 17cm 4cm},clip]{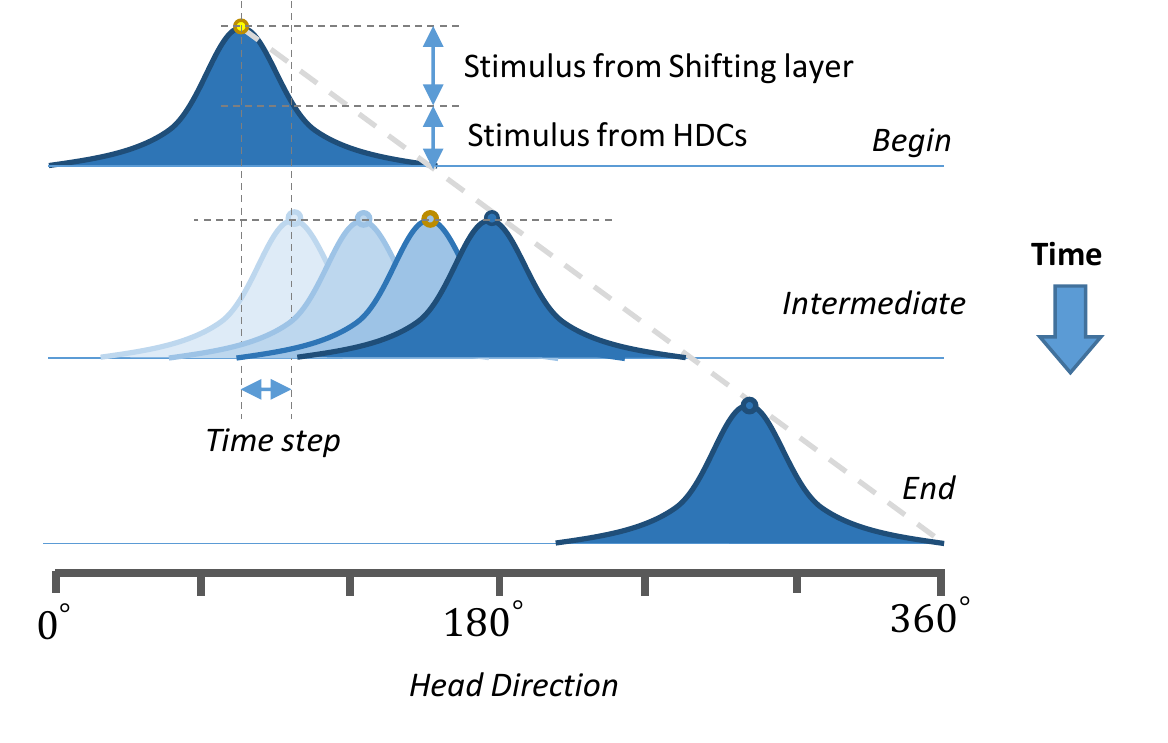}
	\caption{\textcolor{black}{Illustration of the shift of the peak activity in the HDC layer \cite{zhang1996representation}.
			The curve with solid color represents the current head direction over time. The curves with shaded color mean the shifting process of the peak activity.}}
	\label{fig:shiftting}
\end{figure}

\section{Continuous Shifting Mechanism}
In this section, we present the continuous shifting mechanism to shift the activity peak of the HDC layer with an angular velocity given to the network in the form of external stimuli. 

\subsection{Continuous Shifting Mechanism}
The principle for shifting the peak activity of the HDC layer is to add external stimuli to all the HDCs to drive them towards the same direction while keeping the overall activity profile of the HDC layer. 
The activity of the HDC layer is designed to be internally stable, thus we need to add additional shift layers to perform such shifting behaviors in the form of external cues, such as the self-estimated angular velocity of the agent.
We design two shift layers (towards left and right) that are parallel to the HDC layer and each HDC has one corresponding cell in the shift left layer and one in the shift right layer (See Figure~\ref{fig:topology}).
Therefore, cells in the shift layer will be referred to with the same numbering system used for the HDCs.

An external stimulus to one of the shift layers should cause a shift in the HDC network's activity peak (See Figure~\ref{fig:shiftting}). 
Thus, each cell in a shift layer has excitatory synaptic connections to HDCs in the shift direction and inhibitory connections to HDCs in the other direction. 
The shift layers have no impact on the activities of the HDC network when there is no external stimulus. 
If both layers receive the same stimulus, the shift left layer and shift right layer to the HDC layer will cancel out each other. 

\textcolor{black}{
	The dynamic activity shift of the HDC layer occurs when the synaptic weight distribution $w(\Delta \theta, t)$ has a non-zero component.
	To ensure a shift of the HDC's activity profile, the required weight distribution function can be defined as
	\begin{equation}
		\begin{split}
			w(\Delta \theta, t) =W(\Delta \theta) + \gamma(t) W'(\Delta \theta) &  \approx W(\Delta \theta + {\gamma(t)}) \text{,}
		\end{split}
		\label{fig_shift_w}
	\end{equation}
	where $W$ is the same weight distribution and $W'$ is the derivative.
	$\gamma(t)$ is a time-varying coefficient.
	The net effect of adding the derivative $W'$ is approximately a shift of
	the original weight distribution $W$ by an angle equal to the value of $\gamma (t)$.
	Due to the nature of the convolution $U = W * F$, we can infer that $U$ is shift-invariant when $W$ is shift-invariant by following \eqref{fig_shift_w}.
	As long as the weights from the shift layer to the HDC layer meets the form $\gamma(t) W'(\theta)$, the activity profile can be ensured to shift without any shape distortion.
}
\textcolor{black}{
	Thus, the synaptic weight $w_{left}^{S \rightarrow H}$ from the shift left layer to the HDC layer is defined as:
	\begin{equation}
		w_{left}^{S \rightarrow H} = \gamma(t) W'(\Delta \theta) \text{,} 
		\label{equ:shift_weight}
	\end{equation}
	We further define 
	\begin{equation}
		w_{right}^{S \rightarrow H} = -w_{left}^{S \rightarrow H} \text{.}
		\label{equ:shift_weight_right}
	\end{equation}
	The synaptic weights connecting the shift layer and the HDC layer are also visualized with the dotted line and the dashed dot line in Figure~\ref{fig:all_weights}.}

%On the basis of \eqref{equ:shift_weight} and \eqref{equ:shift_weight_right}, we can obtain
%\begin{equation}
%\begin{split}
%W(\Delta \theta) + \gamma(t) W'(\Delta \theta) &  \approx W(\Delta \theta + {\gamma(t)}) \text{.}
%\end{split}
%\label{fig_shift_w}
%\end{equation}
%\textcolor{black}{
%The net effect of adding the derivative $W'$ is approximately a shift of
%the original weight distribution $W$ by an angle equal to the value of $\gamma (t)$.}
%Due to the nature of the convolution $U = W * F$, we can infer that $U$ is shift-invariant when $W$ is shift-invariant by following \eqref{fig_shift_w}.
%%
%%ensure that as long as the weights from the shift layer to the HDC layer meets the form $\gamma(t) W'(\theta)$, the activity profile can be ensured to shift without any shape distortion, which can be proved by
%%
%Moreover, the instantaneous shift speed of the HDC layer is simply proportional to the magnitude $\gamma (t)$.

\subsection{Continuous Shifting with External Stimuli}
Now, we have to prove that the HDC network is shift-invariant with the present of external stimuli.
Each HDC has two stimulus sources $u^{HDC}$ and $u^{shift}$, which come from the other HDCs and the cells in the shift layer, respectively.
For all stimuli injected to the $i^{th}$ HDC, it can be represented as follows.
\begin{equation}
	\begin{split}
		u^{total} & = u^{HDC} + u^{shift} \\
		&  = \sum_{j=1,~i\neq j}^n \bigg(w^{H \rightarrow H}_{ij}f^{HDC}_j + w^{S \rightarrow H}_{ij}f^{shift}_j \bigg)
	\end{split} \text{,}
	\label{equ:total_shift_weight_1}
\end{equation}
where $w^{H \rightarrow H}$ is the synaptic connection from other HDC cells to the $i^{th}$ HDC and $w^{S \rightarrow H}$ is the synaptic connection from the shift layer to the $i^{th}$ HDC.
$f^{HDC}$ and $f^{shift}$ are the firing rates of the cells in the HDC layer and the shift layer, respectively.
Each cell in the shift layer also receives two stimulus sources, which comes all the HDCs and the turning cell (external input $I^{ext}$).
Then, we can obtain
\begin{equation}
	\begin{split}
		u^{shift} & =  \sum_{j=1,~i\neq j}^n \bigg(w^{S \rightarrow H}_{ij}f^{shift}_j \bigg)\\
		&  = \sum_{j=1,~i\neq j}^n \bigg(w^{S \rightarrow H}_{ij} \phi( w^{H \rightarrow S}_{ij} f^{HDC}_j + I^{ext})  \bigg)\\
	\end{split} \text{,}
	\label{equ:total_shift_weight_2}
\end{equation} 
where $w^{H \rightarrow S}$ is the synaptic connection from the HDC layer to the $i^{th}$ shift cell.
Now we can rewrite \eqref{equ:total_shift_weight_1} as
\begin{equation}
	\begin{split}
		u^{total} = \sum_{j=1,~i\neq j}^n \bigg(w^{H \rightarrow H}_{ij}f^{HDC}_j + w^{S \rightarrow H}_{ij} \phi( w^{H \rightarrow S}_{ij} f^{HDC}_j + I^{ext})  \bigg)\\
	\end{split} \text{,}
	\label{equ:total_shift_weight}
\end{equation}

The activity profiles of both shifting layers are designed to have the same shape as the HDC layer's activity profile but half of the amplitude, by setting $w^{H \rightarrow S} = \frac{1}{2}w^{H \rightarrow H}$ (See the dashed line in Figure~\ref{fig:all_weights}). 
The reason for setting a lower amplitude for the shifting layers is to leave space for the injection of the external stimulus to the shift cells.  
Thus, a cell in one of the shift layers receives half of the synaptic inputs from the HDC layer as the corresponding HDC receives from the HDC layer.

\begin{figure}[!tpb]
	\centering
	\includegraphics[width=0.45\textwidth]{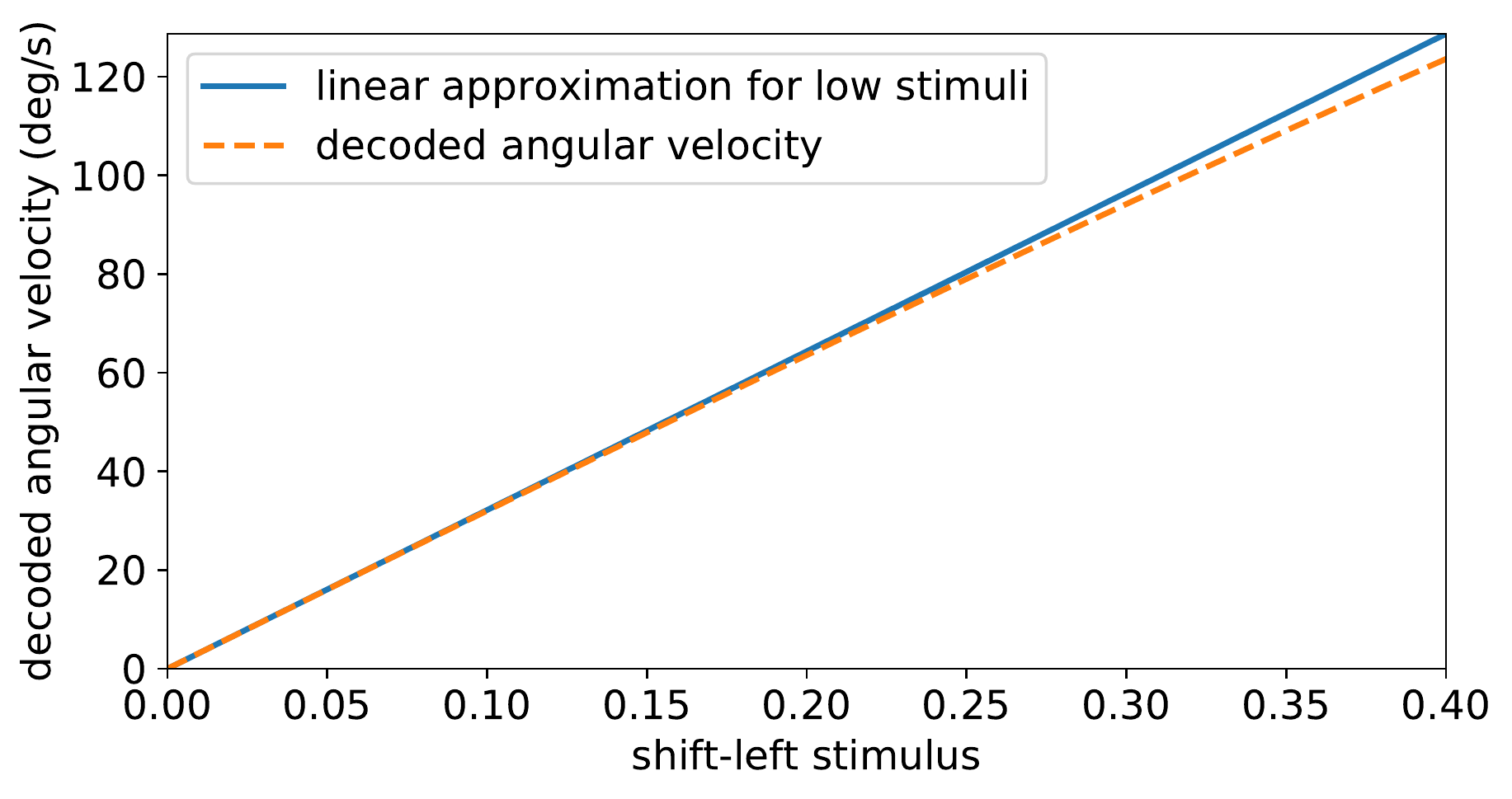}
	\caption{Angular velocity of the activity peak for different stimuli to the shift-left layer.}
	\label{fig:velocity_stim}
\end{figure}

On the one hand, we assume $I^{ext} = \alpha v$, where $v$ is the angular velocity of the robot and $\alpha$ is a constant coefficient.
Then, the firing rate of the cell in the shift layer stimulated by the HDC layer and external stimuli can be calculated as 
\begin{equation}
	\begin{split}
		\phi\big(\sum_{j=1,~i\neq j}^n w^{H \rightarrow S}_{ij}f^{HDC}_j + I^{ext}\big) &= \phi \big( \sum_{j=1,~i\neq j}^n \frac{1}{2} w^{H \rightarrow H}_{ij}f^{HDC}_j + \alpha v\big) \\
	\end{split}
	\label{equ:external_stimulus}
\end{equation}
On the other hand, according to \eqref{eq:phi_neuron}, the firing rate is approximately linear to its input when the input is in the range of $(0, 4)$ (See the shadow area in Figure~\ref{fig:phi}). 
Then we can obtain:
\begin{equation}
	\underbrace{\phi \big( \sum_{j=1,~i\neq j}^n \frac{1}{2} w^{H \rightarrow H}_{ij}f^{HDC}_j + \alpha v\big)}_{f^{shift}} - \underbrace{\phi \big( \sum_{j=1,~i\neq j}^n \frac{1}{2} w^{H \rightarrow H}_{ij}f^{HDC}_j)}_{\frac{1}{2}f^{HDC}} \propto \alpha v \text{.}
\end{equation}
Then we can get
\begin{equation}
	f^{shift} = \frac{1}{2}f^{HDC} + \frac{1}{2}f^{HDC} K \alpha v \text{,}
	\label{equ:proportion}
\end{equation}
where $K$ is a constant coefficient determined by \eqref{eq:time_evolution_neuron}.
A brief mathematical proof of \eqref{equ:proportion} can be found in Appendix \ref{appendix_2}.
Then, we can rewrite \eqref{equ:total_shift_weight_1} as
\begin{equation}
	u^{total} = \sum_{j=1...n,~i\neq j} \bigg(w^{H \rightarrow H}_{ij}f^{HDC}_j + \frac{1}{2}(1 + K \alpha v) w^{S \rightarrow H}_{ij} f^{HDC}_j \bigg)
\end{equation}
Finally, according to \eqref{equ:shift_weight}, by choosing $w^{S \rightarrow H}_{ij}$ to meet $(1 + K \alpha v)  w^{S \rightarrow H}_{ij} \propto W'(\Delta \theta)$, we can ensure the peak activity of the HDC layer can shift without shape distortion.

To find a proper coefficient between the external turning stimulus and the shift speed of the peak activity of the HDC layer, we run a series of numerical simulations by applying a range of stimuli for turning left and recording the corresponding angular velocity decoded by the HDC network.
Then, we can fit out the relationship between the turning speed of the agent and the decoded turning speed of the HDC network.
The results are shown in Figure~\ref{fig:velocity_stim}.
By inverting that linear approximation, a function mapping the angular velocity to the stimulus is obtained. It is given as:
\begin{equation}
	I_s=0.178124v \text{,}
	\label{eq:stim_av}
\end{equation}
where $I_s$ is the stimulus to cells in the the shift layer corresponding to the turning direction and $v$ is the absolute value of the angular velocity given in radians per second.

\textcolor{black}{It should be noted that the potential noises that are introduced in the shit mechanism are mainly from two sources, namely, the stochastic neural activity of a neuron and the external stimulus to the shift layers.
	In this paper, since we use deterministic non-spiking neuron, the neural activity will not bring any noise error.
	For a more refined spiking-based neuron model, a more biologically plausible way is to model the neural activity using neuron population, in which the averaged firing rate is used to represent the neuron population.
	For the noise errors that are introduced by sensory information, we will provide experiment results in the next section.}

%The connections of synapses between HDC layer and shift layers are also visualized in Figure~\ref{fig:topology}.
%Each neuron in the HDC layer is connected to its corresponding neuron in the shift-left and shift-right neuron to keep them activated.
%The activity profile of both shifting layers is designed to have the same shape as the HDC layer's activity profile but with a lower amplitude. 
%The reason for setting a lower amplitude for the shifting layers is to leave spaces for the injections of the external stimulus from the \hl{rotation cells}.  
%Connections from the HDC layer to a neuron in one of the shift layers are defined as the connections from the HDC layer to one HDC scaled with $\frac{1}{2}$, i.e. by the weight function shown in Figure \ref{fig:all_weights} scaled with $\frac{1}{2}$. 
%Thus, a neuron in one of the shift layers receives half the synaptic inputs from the HDC layer as the corresponding HDC receives from the HDC layer.

%On one hand, in order to shift the activity peak smoothly and without shape distortion, \textcolor{black}{we have to ensure that the signal intensity of the shift stimulus is strong enough to fire the neighbouring HDCs.}  
%On the other hand, the stimulus from the shifting layer should excite the activities of the HDC layer to shift to one direction and inhibit the activities from shifting to the other direction.

\section{Experiments}
\label{sec_experiment}

\begin{figure}[!tb]
	\centering
	\includegraphics[width=0.48\textwidth]{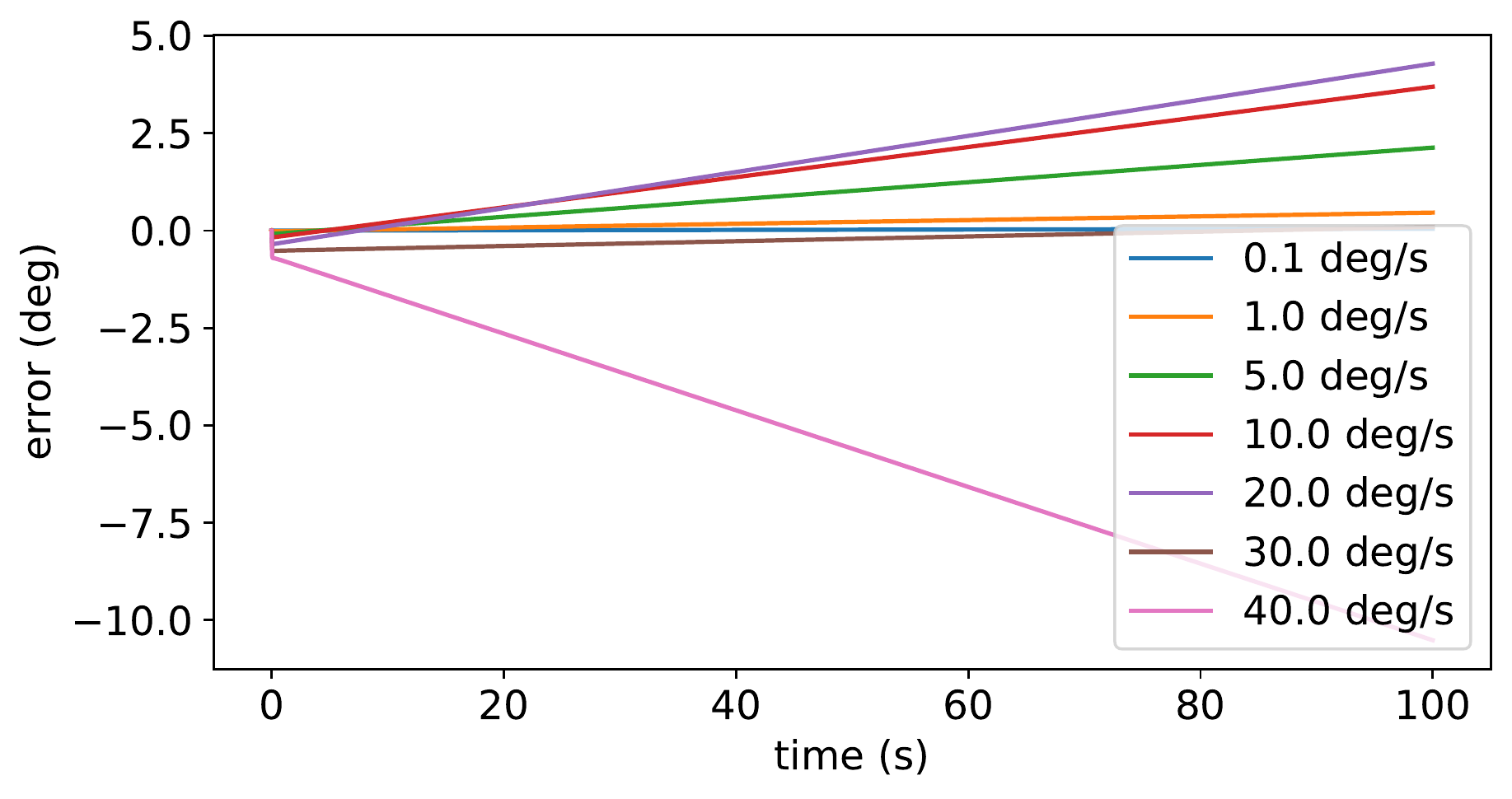}
	\caption{The accumulated errors under one-direction circular motion with different angular velocities.}
	\label{fig:error_circle}
\end{figure}

In this section, we will evaluate the performance of our proposed HDC network in both simulations and real-world implementations. 
We first run a series of numerical simulations to show the tracking accuracy of the directional heading with different angular velocities.
We then test the HDC network in a challenging maze navigation task.
We finally present the results of offline testing with the real-world dataset KITTI and the online testing with indoor navigation experiments.

\subsection{Simulations}

\subsubsection{Numerical Simulation}
First, we run a series of numerical simulations to show the tracking accuracy under one-direction circular motion with different angular velocities.
Since the robot turns to one direction all the time, the tracking error will be accumulated over time.
From Figure~\ref{fig:error_circle}, we can find that the errors steadily accumulate over time, but are limited in a small range even running for a long time.
When the angular velocity is below $40^{\circ}$/s, the averaged error against a full lap ($360^{\circ}$) is less than $1^{\circ}$.
%This indicates, a calibration mechanism with external cues, such as visual land marks, is necessary to keep accuracy for longer runs.

\begin{figure}[!t]
	\centering
	\includegraphics[width=0.42\textwidth, trim={1cm 12cm 1cm 1cm},clip]{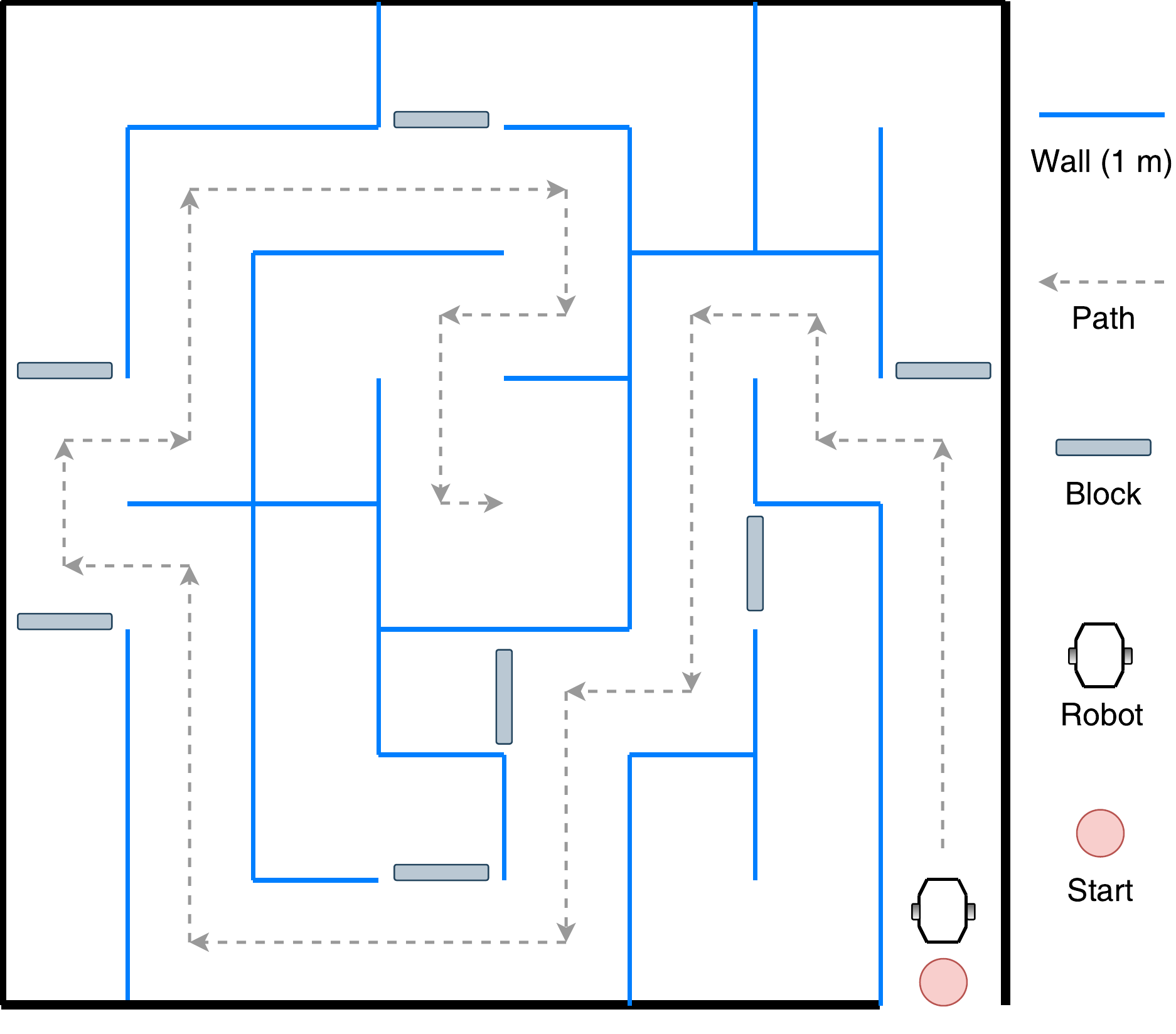}
	\caption{The schematic diagram of the maze environment.}
	\label{fig:maze_env}
\end{figure}

\subsubsection{Robot Simulation}
To provide simple and configurable environments to test the HDC network, we set up a simulated robotic navigation task using PyBullet~\cite{pybullet} \footnote{The simulation can be viewed at \href{https://videoviewsite.wixsite.com/biologicalhdc}{https://videoviewsite.wixsite.com/biologicalhdc}}.
As shown in Figure~\ref{fig:maze_env}, the maze-like environment consists of a collections of walls and only one path will lead the robot to the center of the maze, which requires similar numbers of left and right turns.
A Pioneer robot~\cite{zaman2011ros} is placed at the bottom right of the maze and takes the direct path into the middle of the maze, turns around and drives the same way back.
The robot is equipped with 16 proximity sensors to detect any obstacles and controlled by a Braitenberg controller which allows it to avoid obstacles and navigate freely inside the maze~\cite{french2005introducing}.
The robot’s angular velocity is injected into the HDC network as external stimuli according to \eqref{eq:stim_av}.
It should be noted that, for instance, the $v_{right}$ is set as zero when the robot turns left.

The HDC network is simulated with a step size of $0.5$ ms inspired by \cite{NEURONAttractor}. 
The Pybullet has a default simulation timestep of $50$ ms.
Since the simulation step for the neural network is significantly faster than a PyBullet simulation step, the HDC network is simulated for $100$ timesteps without change in stimuli during every PyBullet timestep. 
To verify that the higher PyBullet timestep doesn't impact accuracy, the simulation was also done with a PyBullet timestep of $25$ ms, showing no significant differences. 
The simulation runs approximately seven times as fast as the real time on the test hardware (AMD Ryzen 5 2600, Nvidia GeForce GTX 1070).

\begin{figure}[!t]
	\centering
	\begin{subfigure}{.48\textwidth}
		\centering
		\includegraphics[width=1\textwidth]{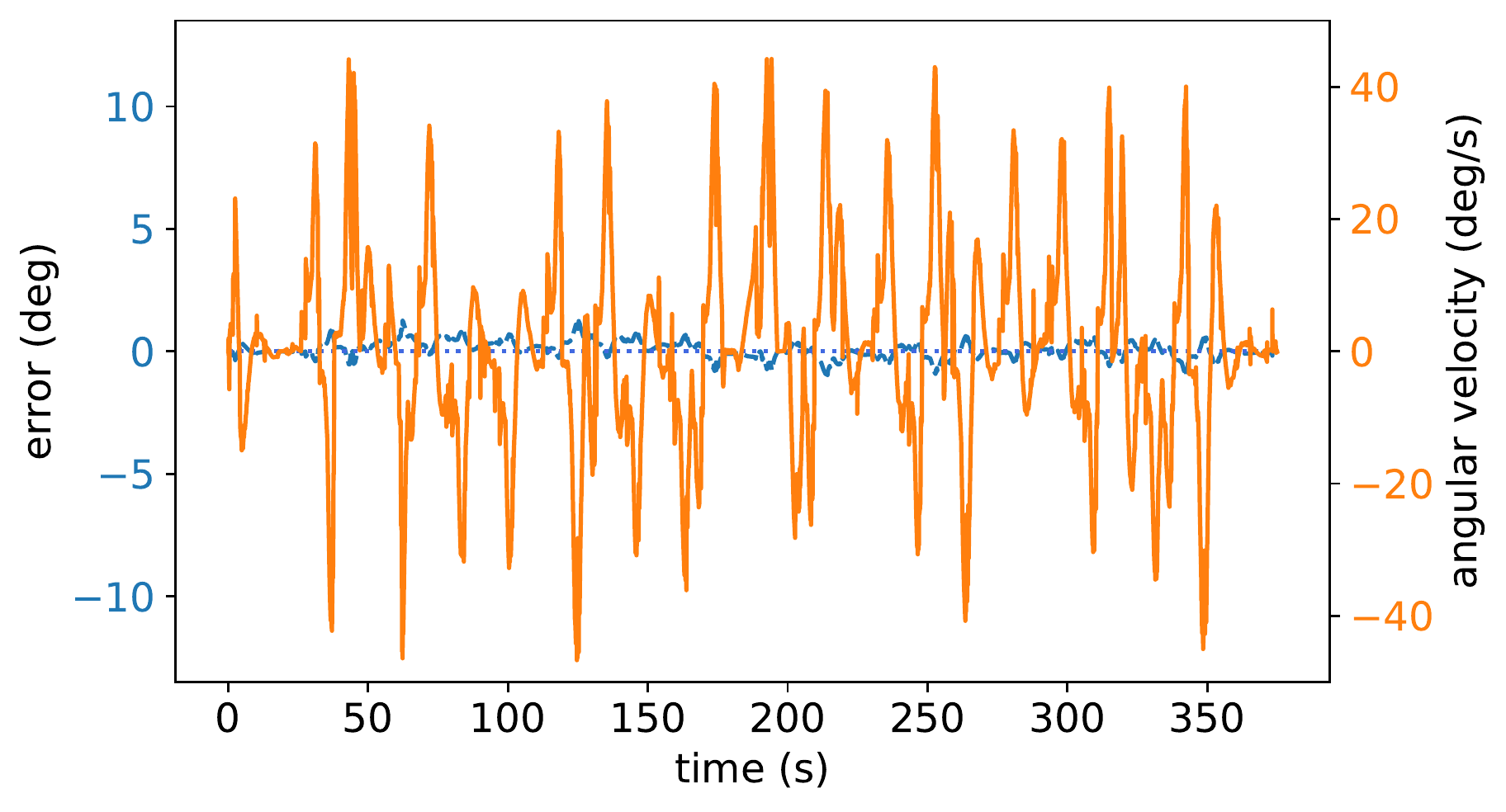}
		\caption{Tracking error and the corresponding angular velocity.}
		\label{fig:err_vs_av}
	\end{subfigure}
	\begin{subfigure}{.45\textwidth}
		\centering
		\includegraphics[width=1\textwidth]{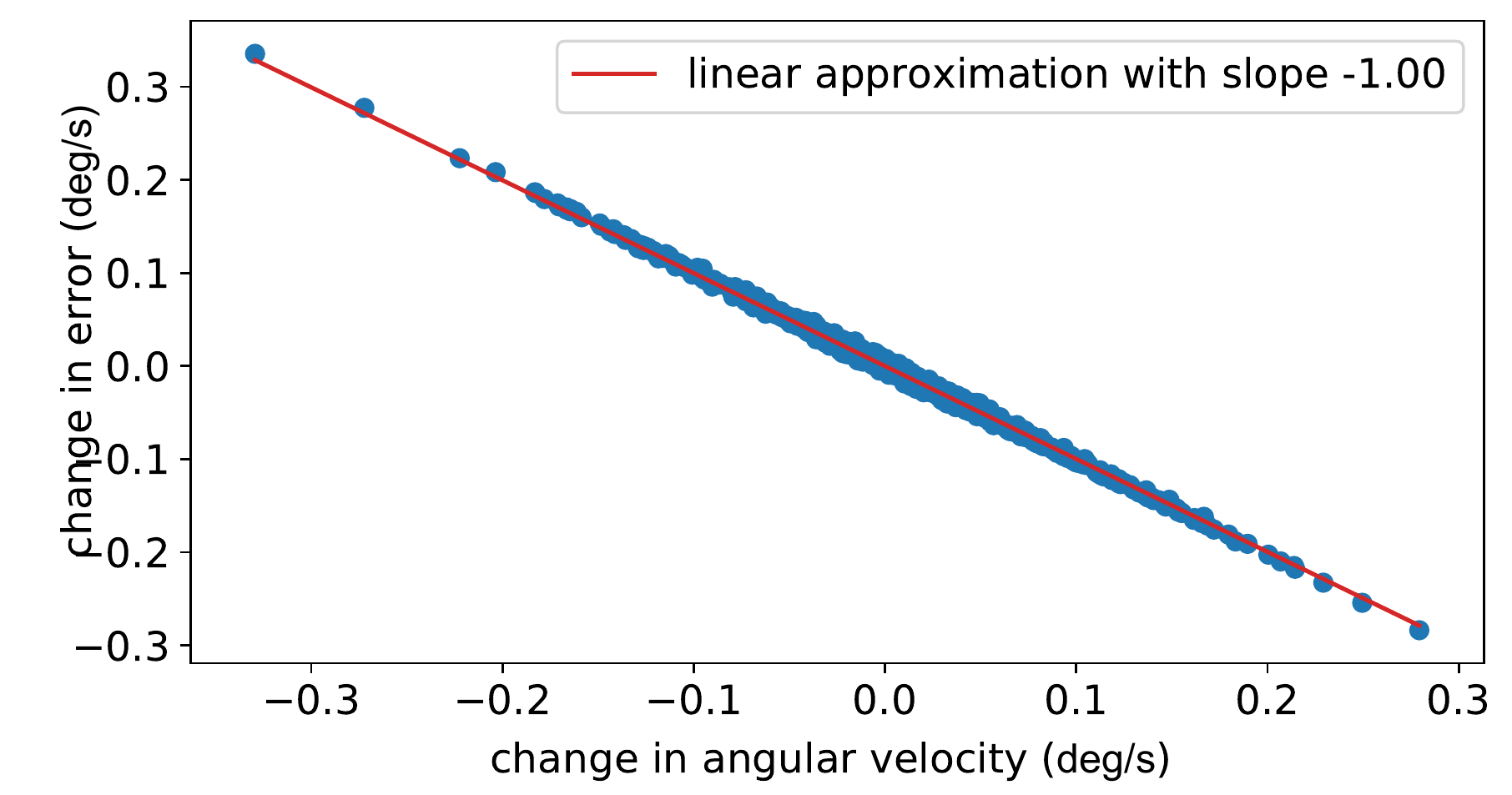}
		\caption{Change in angular velocity vs. change in the error.}
		\label{fig:av_maze}
	\end{subfigure}
	\caption{Error between the direction decoded from the HDC network and the true orientation of the robot over the full simulation episode in the maze environment. The error stays below $1.5^{\circ}$ during the entire simulation.}
	\label{fig:result_maze}
\end{figure}
\begin{figure*}[!t]
	\centering
	\begin{subfigure}{0.85\textwidth}
		\centering
		\includegraphics[width=1\textwidth]{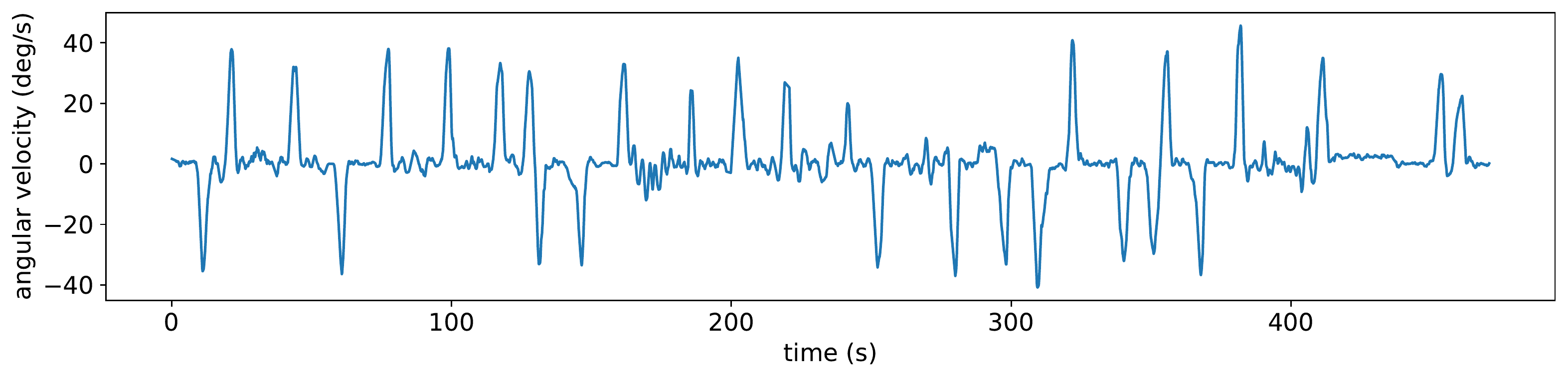}
		\\
		[-1.5ex]
		\caption{The angular velocity.}
		\label{fig:angular_velocity_KITTI}
	\end{subfigure}
	\begin{subfigure}{0.85\textwidth}
		\centering
		\includegraphics[width=1\textwidth, trim={0.0cm 5.5cm 0.1cm 5.5cm},clip]{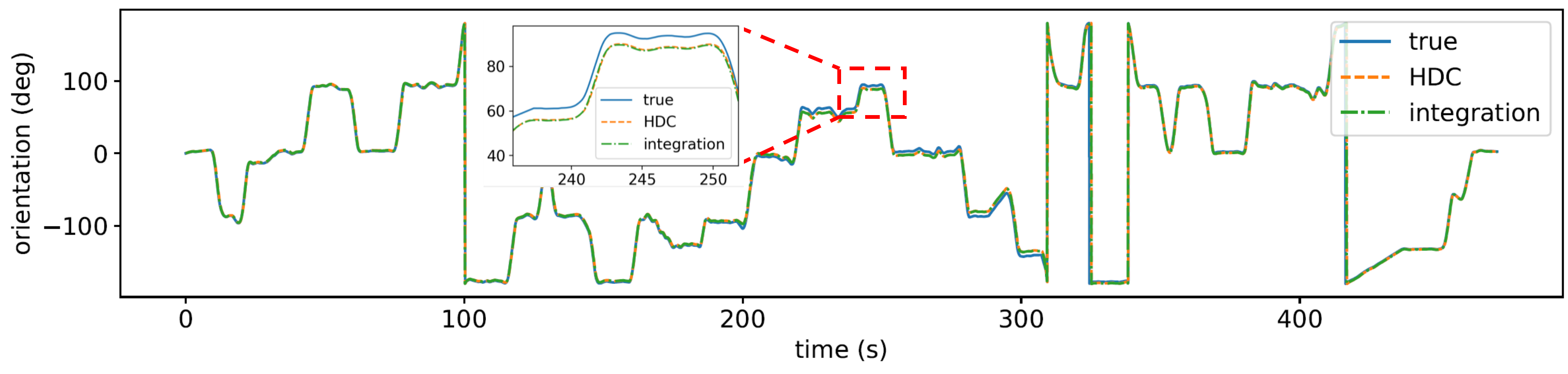}
		\\
		[-1.5ex]
		\caption{The orientation data (directional heading) of the ground truth, the HDC, and the numerical integration.}
		\label{fig:orientation_KITTI}
	\end{subfigure}\\
	\caption{The experimental results of the scenario\textit{ avs\_kitti\_raw\_2011\_10\_03\_drive\_0027}.}
	\label{fig:KITTI}
\end{figure*}

The experiment results of the maze-like environment are shown in Figure~\ref{fig:result_maze}.
The estimation error, defined as the difference between the direction decoded from the HDC network and the ground truth, is shown in Figure~\ref{fig:err_vs_av}.
The angular velocity of the agent is also recorded through the whole simulation.
We can find that, even the agent changes its direction significantly at a maximum turning speed $40^{\circ}$/s, the error is still maintained below $\pm1.5^{\circ}$.
During the whole simulation, the changing direction of the robot accumulated through the simulation is around $4000^{\circ}$ over both left and right turns.
To have a close look at Figure~\ref{fig:err_vs_av}, we can also find that the estimation error often moves opposite to the angular velocity. 
This indicates a slight lag in the HDC network's response, which is expected since cells don't respond instantly to stimuli.
From Figure~\ref{fig:av_maze}, we can find the tracking error is symmetrically aligned with the turning direction, which also explains why the error has not been accumulated when the robot turns left or right at a similar number of times.

%\begin{figure*}[!t]
%	\centering
%	\includegraphics[width=0.85\textwidth]{python_plot/error_average_KITTI.pdf}
%	\\
%	[-1ex]
%	\caption{The difference between the integration error and HDC network error is plotted as a gray line for each of the 156 KITTI scenario. A negative value means that the HDC network performed better than integration.}
%	\label{fig:error_kitti}
%\end{figure*}

\subsection{Offline Real-world Tests}
To test the performance of the HDC network on an automobile in urban scenarios, we first present the offline testing results on the KITTI dataset \cite{kitti}. 

The KITTI dataset consists of recordings from a car equipped with multiple sensors including an Inertial Measurement Unit (IMU). 
From the KITTI dataset, we take the orientation data recorded by the IMU as the ground truth and the angular velocities as the input for the HDC network. 
\textcolor{black}{
	The angular velocity data is recorded with noisy, which affects the accuracy of the decoded directional heading from the HDC network.}
Since the HDC network is expected to perform integration of the angular velocity, a proper benchmark will be the numerical integration of the angular velocities with the trapezoid rule, i.e. approximating each time interval of $100$ms with a trapezoid.
Taking the scenario \textit{avs\_kitti\_raw\_2011\_10\_03\_drive\_0027} as an example, the directional headings from the ground truth, the benchmark, and the the HDC network are compared in Figure~\ref{fig:KITTI}, together with the angular velocity data. 
The reason for choosing this scenario is because it shows the highest error among all the $156$ KITTI scenarios.
On the one hand, compared with the ground truth, the averaged error over time is around $2.46^{\circ}$
and the maximum error is less than $12^{\circ}$.
On the other hand, compared with the numerical integration of the angular velocity, the averaged error is around $1.11^{\circ}$ and the maximum error is less than $3.29^{\circ}$.
Finally, we present the errors of the HDC network against the ground truth over all the KITTI scenarios that have a time duration longer than $10$ s (See Table \ref{tab_all_results} in the Appendix). 
From this table, we can find that the averaged errors are limited below $3^{\circ}$ for all the scenarios.
The maximum errors are also less than $6^{\circ}$, except for the scenario \textit{avs\_kitti\_raw\_2011\_10\_03\_drive\_0027}. 

\subsection{Online Read-World Tests}

To examine the real-time capability of the proposed HDC network and its performance on MCUs, we second run an online experiment by performing an indoor robotic navigation task.

\begin{figure*}[!tb]
	\centering
	\begin{subfigure}{0.48\textwidth}
		\includegraphics[width=1\textwidth]{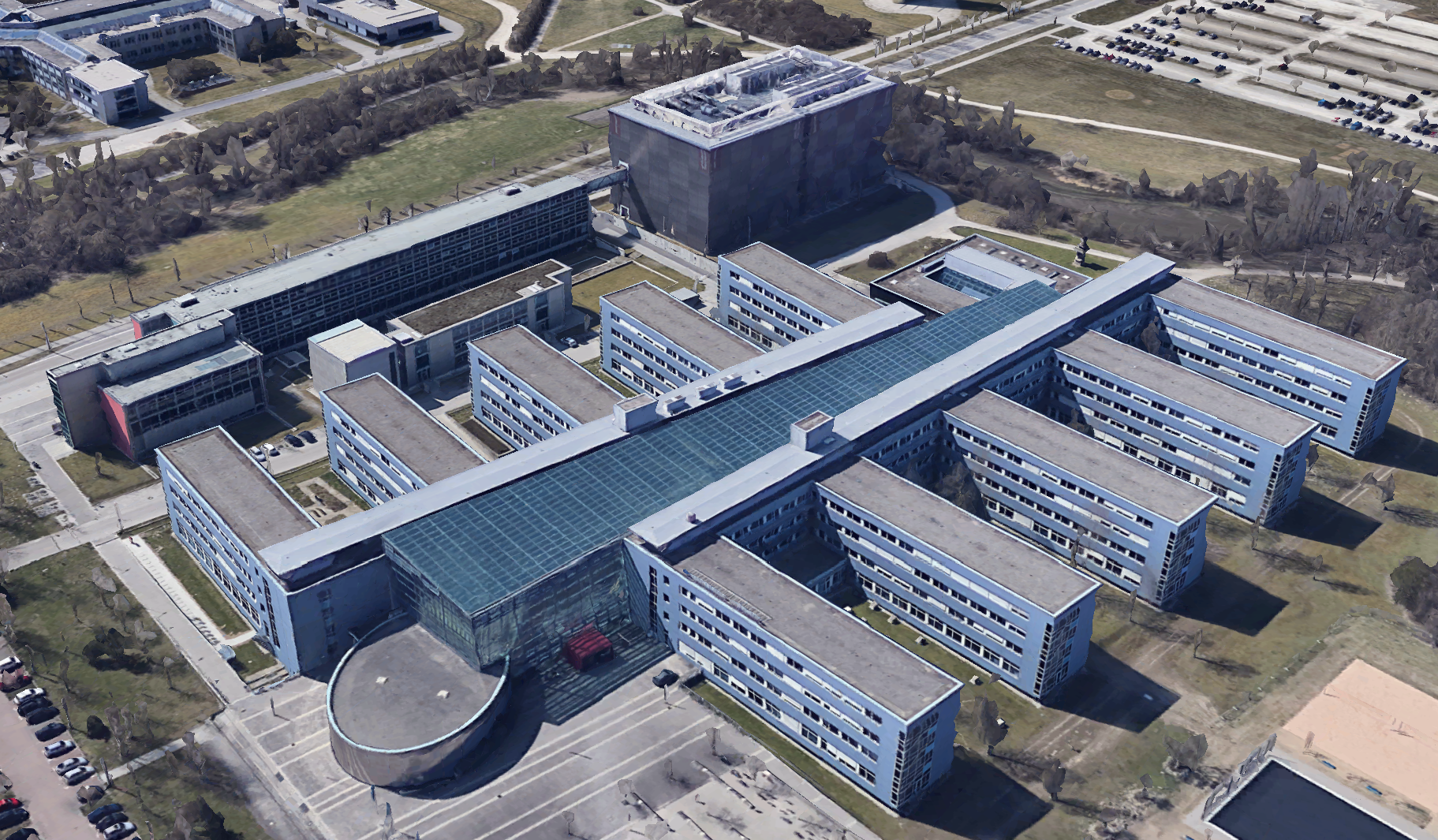}
		\caption{The floor map of the FMI, TUM.}
		\label{fig:floor_plan}
	\end{subfigure}
	\begin{subfigure}{0.48\textwidth}
		\includegraphics[width=1\textwidth, trim={0.0cm 0.0cm 0.0cm 0.0cm},clip]{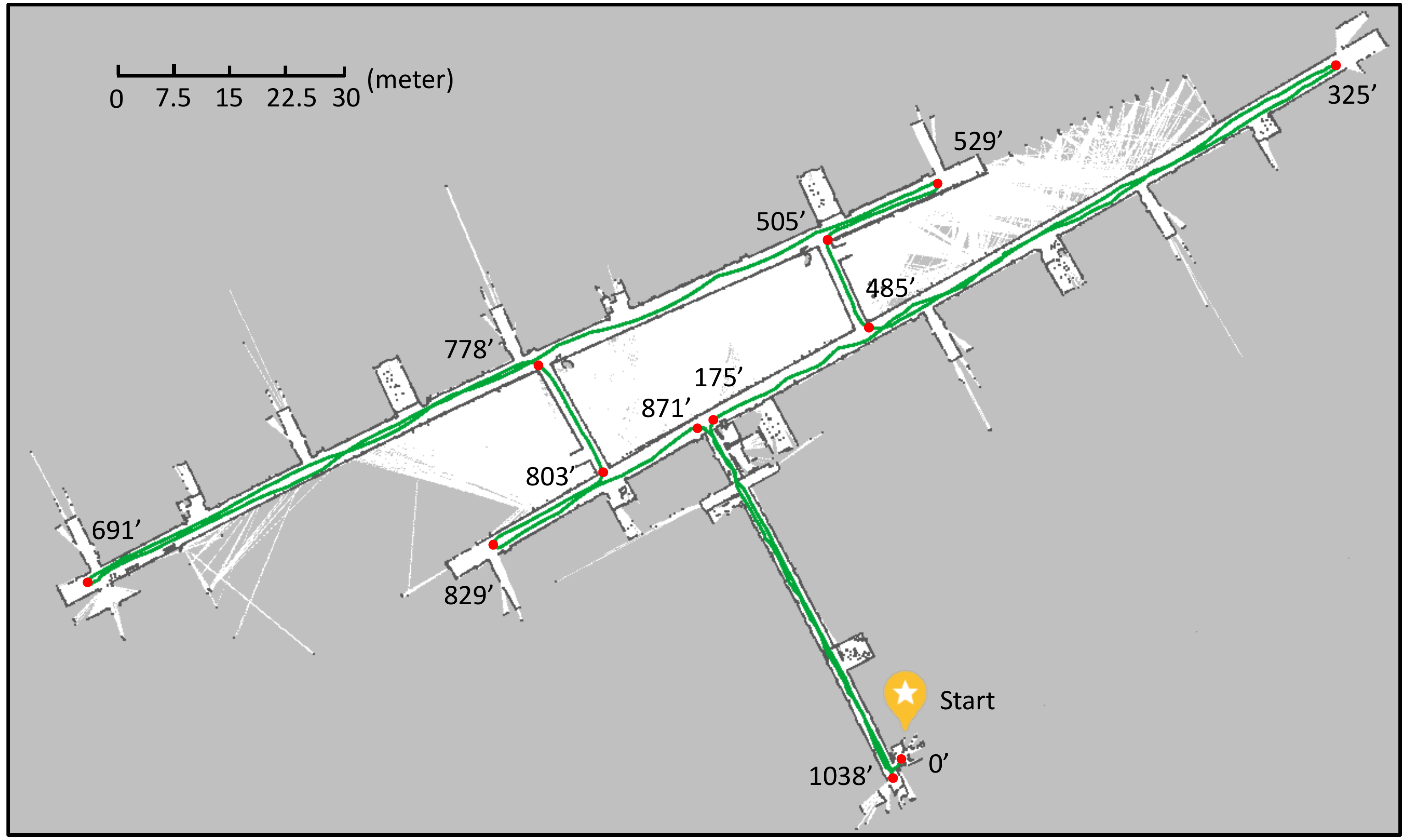}
		\caption{The trajectory of the robot in the FMI scenario.}
		\label{fig:trajectory}
	\end{subfigure}
	\caption{The floor map of the navigation scenario and the trajectory of the mobile robot.}
\end{figure*}

The mobile robot platform is a remotely controlled model racing car, which is equipped with a Raspberry Pi, a camera, a Lidar sensor, and an IMU sensor.
The robot was manually operated to navigate in the building of the Faculty of Mathematics and Informatics (FMI), Technical University of Munich.
The floor map of the FMI building is shown in Figure~\ref{fig:floor_plan}.
The trajectory of the robot is calculated using the dead reckoning method based on the Lidar odometry~\cite{kohlbrecher2011flexible} and visualized in Figure~\ref{fig:trajectory}, with the timestamp at each turning corner.
The car started from the office at $0$s, navigated along the corridors of the building, and finished the lap at $1038$s.
The averaged speed of the car was about $0.5$ m/s through the experiment. 

\begin{figure*}[t]
	\centering
	\begin{subfigure}{0.85\textwidth}
		\centering
		\includegraphics[width=1\textwidth]{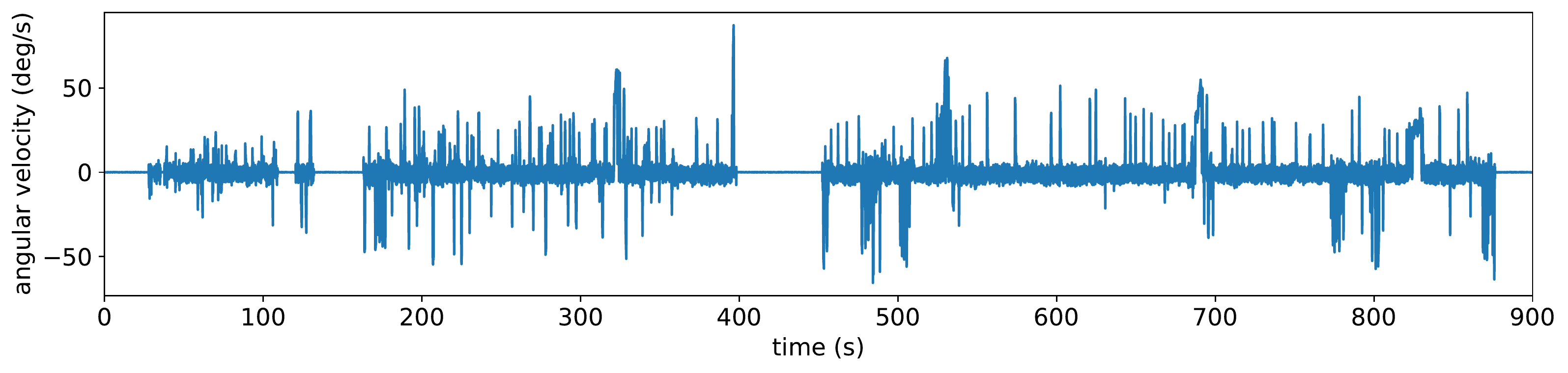}
		%		\\[-2ex]
		\caption{The angular velocity.}
		\label{fig:angular_velocity_FMI}
	\end{subfigure}
	\begin{subfigure}{0.85\textwidth}
		\centering
		\includegraphics[width=1\textwidth, trim={0.0cm 4cm 0.0cm 1.2cm},clip]{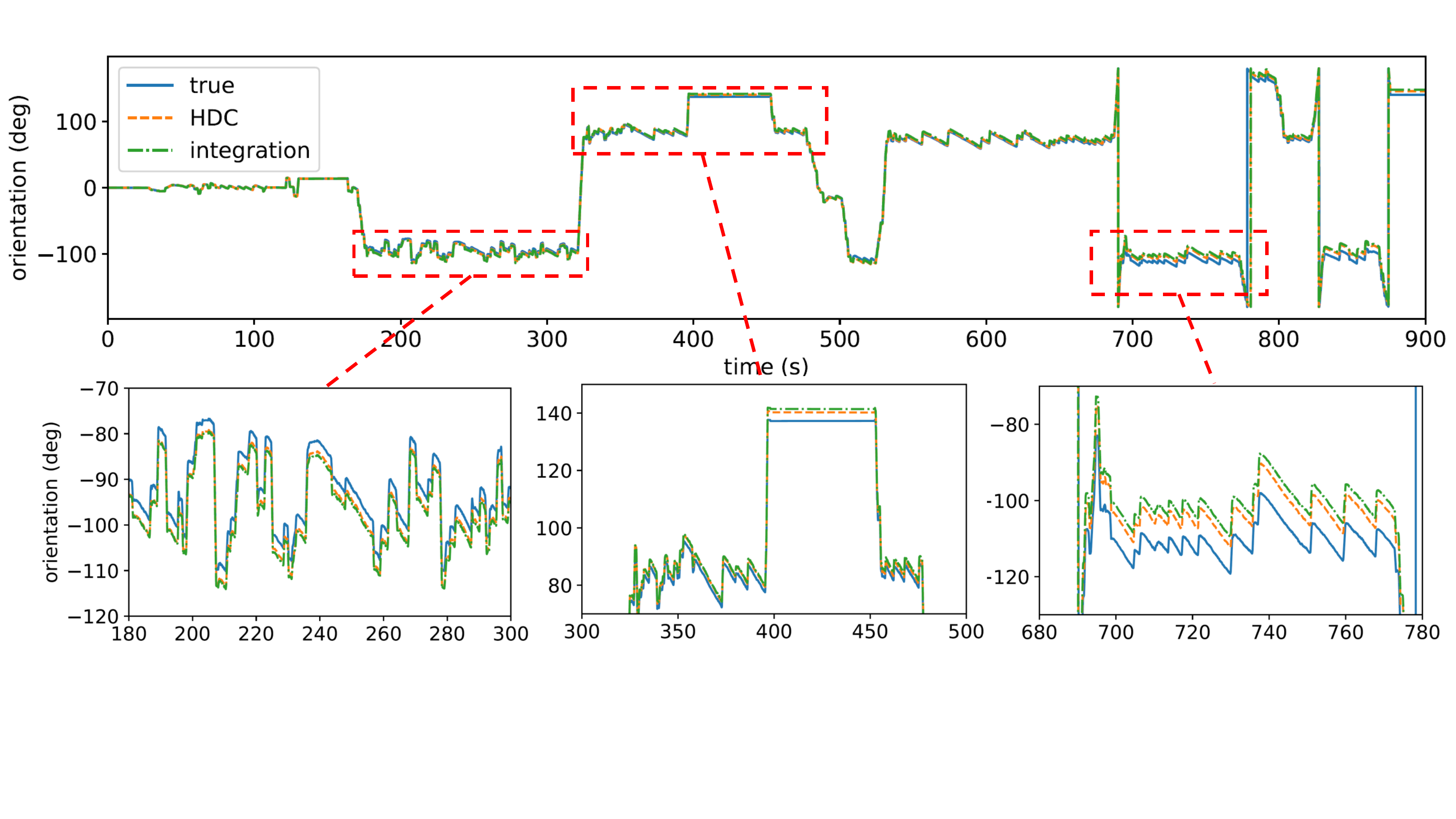}
		\caption{The orientation data (directional heading) of the ground truth, the HDC, and the numerical integration.}
		\label{fig:orientation_FMI}
	\end{subfigure}\\
	\caption{The experimental results of the FMI scenario.}
	\label{fig:error_private}
\end{figure*}

The ground truth of the directional heading and the angular velocity were recorded by the on-board IMU.
%We took the orientation data as the ground truth and the angular velocity as the input for the HDC network.
Figure~\ref{fig:error_private} shows the results of the online testing during the time between $0$s to $900$s in the FMI building.
The angular velocity is shown in Figure~\ref{fig:angular_velocity_FMI}.
Since we used a cheap electronic IMU sensor (BNO055), the angular velocity data was more noisy than the data in the KITTI dataset.
There are several line segments without noise that correspond to the time when the car was temporarily stopped.


\textcolor{black}{
	This noisy angular velocity leads to a higher error in this experiment, in which the averaged error is $3.4^{\circ}$ and the maximum error is $10.07^{\circ}$ for the numerical integration method based on the trapezoid rule.
	For the HDC network, the averaged error is around $1.11^{\circ}$ and the maximum error is less than $3.69^{\circ}$.}1
On the one hand, compared with the results of the KITTI, the reason for higher errors is mainly caused by the noisy of the angular velocity data.
On the other hand, we found that the results from the HDC network (orange dash line) are more close to the ground truth (blue solid line) than the results from the numerical integration (green dot dash line), as shown in Figure~\ref{fig:orientation_FMI}.
This is because the dynamics of the HDC network makes it less sensitive to noise than the numerical integration method. 

Moreover, this experiment was also designed to test the real-time capability of the HDC network, which was executed on a Raspberry Pi 3 Model B+.
Since the input data from an IMU sensor was sampled at a frequency of $100$ Hz, the HDC was also executed at $100$ Hz on the Pi. 
We list the computation time in Table~\ref{table_run_time}.
The median running time for processing each frame of data is around $7.7$ ms. 
Although a small number of steps cost more than $10$ ms, the total averaged time $7.70$ ms is still fast enough to guarantee the real-time computing.
Based on these results, we can demonstrate the real-time capability of the HDC network even on a mobile microchip such as the Raspberry Pi.

\begin{table}[!b]
	\renewcommand{\arraystretch}{1.0}
	\caption{Computation time on a Raspberry Pi 3.}
	\centering
	\label{table_run_time}
	\begin{tabular}{p{0.2\textwidth}p{0.1\textwidth}}
		\hline\hline \\[-3mm]
		\multicolumn{1}{l}{Parameters} &  \multicolumn{1}{l}{Quantity} \\ [1.0ex] \hline 
		Averaged time & $7.70$ ms \\
		Median time & $7.43$ ms \\
		Maximum time & $85.66$ ms \\
		Percetange of steps with higher time than $10$ ms & $4.79\%$ \\
		\hline\hline
	\end{tabular}
\end{table}

\section{Biological Plausibility}
To design a biologically plausible HDC network, both the properties and functions of the HDC network should exhibit similar behaviors as HDCs recorded in nature~\cite{taube2007head}. 
\textcolor{black}{
	From the perspective of functionality, the proposed HDC network can keep an estimate of the directional heading of the agent with a relatively small error while only relying on its angular velocity.} 
This is consistent with biological observations, since biological HDCs are found to remain fully functional even in the dark~\cite{taube1990head}. 
The activity peak of the simulated HDC network resembles a biological HDC's tuning curve, as described in Section \ref{sec_model}. 
Both curves roughly follow the shape of a Gaussian bell curve. 
The maximum firing rate of the peak activity of one HDC is set as $76$ Hz, which is supported by experimental data \cite{taube2007head}. 
Thus, the simulated HDC's behavior is relatively consistent with the behavior of biological HDCs. 
The idea of the continuous shifting mechanism is inspired by the vestibular mechanism~\cite{mcnaughton1991dead} and previous work with a focus on robotic applications such as \cite{skaggsModelNeuralBasis1995} and \cite{8967864}. 
The shift-layer cells primarily correlate with the directional heading similar to HDCs, with a peak firing rate of slightly more than $30$ Hz. 
However, unlike the HDCs, they only show activity with changing angular velocities when rotating in one specific direction. 
For example, cells in the shift left layer only increase their activities  during left turns. 
Cells that function like HDCs with a peak firing rate of around $30$ Hz but also correlate with angular velocity were found in rats as reported by \cite{taube2007head}, \cite{taube1995head}. 
However, no such cells that only correlate with the angular velocity when rotating in one specific direction were found in the previous works.
\textcolor{black}{And biological neurons process information using impulses or spikes, while we only use non-spike based neurons.}
\textcolor{black}{
	Moreover, from the perspective of network structure, our HDC network is not biologically inspired but manually designed, which is based on one-dimensional continuous attractor network.
	The one-to-one connection between the shift layer and the HDC layer is also not biologically plausible since most neurons in brains are connected to many other neighboring neurons at the same time.
	There is no clear evidence that the HDCs in animals are structured in such a way.
}

%\begin{figure}[!tbp]
%	\centering
%	\includegraphics[width=0.48\textwidth]{python_plot/error_average.pdf}
%	\caption{The difference between the integration error and HDC network error is plotted as a gray line for each of the $156$ KITTI scenario. This difference will be referred to as error. A negative error means that the HDC network performed better than integration. The blue curve is the average error for all KITTI scenarios still running at a specific time.}
%	\label{fig:error_kitti}
%\end{figure}

\section{Conclusion}
\label{sec_conclusion}

Developing biologically inspired models for mimicking the structure and function of the navigation system in animals is challenging, especially taking the applicability of being used in robotic tasks into consideration.
Aiming at this problem, we propose a biologically inspired HDC network, which is designed to provide the directional heading of the system only relying on its angular velocity.   
%We implement the proposed HDC network in robotic navigation tasks and test the performances in terms of the accuracy and real-time capability.
Our HDC network utilizes both the advantages of experimental data recorded from in-vivo neurons and engineering technologies.
The experiment results demonstrated great performance in terms of accuracy for estimating the directional heading and real-time capability compared with previous studies.
\textcolor{black}{
	For future research, we will continue to investigate a two-dimensional HDC to mimic the structure and functions of grid cells.}

%\input{acknowledgment}
%%\clearpage	
%\appendix
%\label{sec_appendix}
%%\section*{Appendix}

\appendices
\section{}
\label{appendix_1}
The derivation of calculating the weights $W$ is inspired from \cite{zhang1996representation}.
For \eqref{eq:time_evolution_zhang}, the desired stationary state will be
\begin{equation}
	\centering
	u = w * f
	\label{eq_conv}
\end{equation}
In fact, the convolution equation \eqref{eq_conv} is a special case of the \textit{Fredholm integral equations of the first kind}, which is also know to be ill-posed. 
With the concept of regularization, $w$ can be calculated by minimizing the error function:
\begin{equation}
	\begin{split}
		E &= \frac{1}{2 \pi} \int_{0}^{2 \pi} (u-w*f)^2 d\theta + \frac{\lambda}{2 \pi} \int_{0}^{2 \pi} w^2 d\theta \\
		&= \sum_{-\infty}^{\infty} |\hat{u}_n - \hat{w}_n \hat{f}_n |^2 + \lambda \sum_{-\infty}^{\infty} |\hat{w}_n |^2 \text{.}\\
		& = \sum_{-\infty}^{\infty} \big(\hat{u}^2_n - 2\hat{u}_n\hat{w}_n \hat{f}_n + \hat{w}^2_n \hat{f}^2_n + \lambda \hat{w}^2_n  \big)
	\end{split}
	\label{eq_min}
\end{equation}
$\lambda$ is a parameter to control the trade-off between the accuracy and the flatness of the solution.
The solution of $\hat{w}_n$ corresponds to the value that makes $\frac{dE}{d\hat{w}_n} = 0$.
Then $\hat{w}_n$ can be obtained as 
\begin{equation}
	\hat{w}_n = \frac{\hat{u}_n \hat{f}_n}{\lambda + \hat{f}^2_n} \text{.}
\end{equation}

\section{}
\label{appendix_2}
\begin{figure}[!t]
	\centering
	\includegraphics[width=0.3\textwidth]{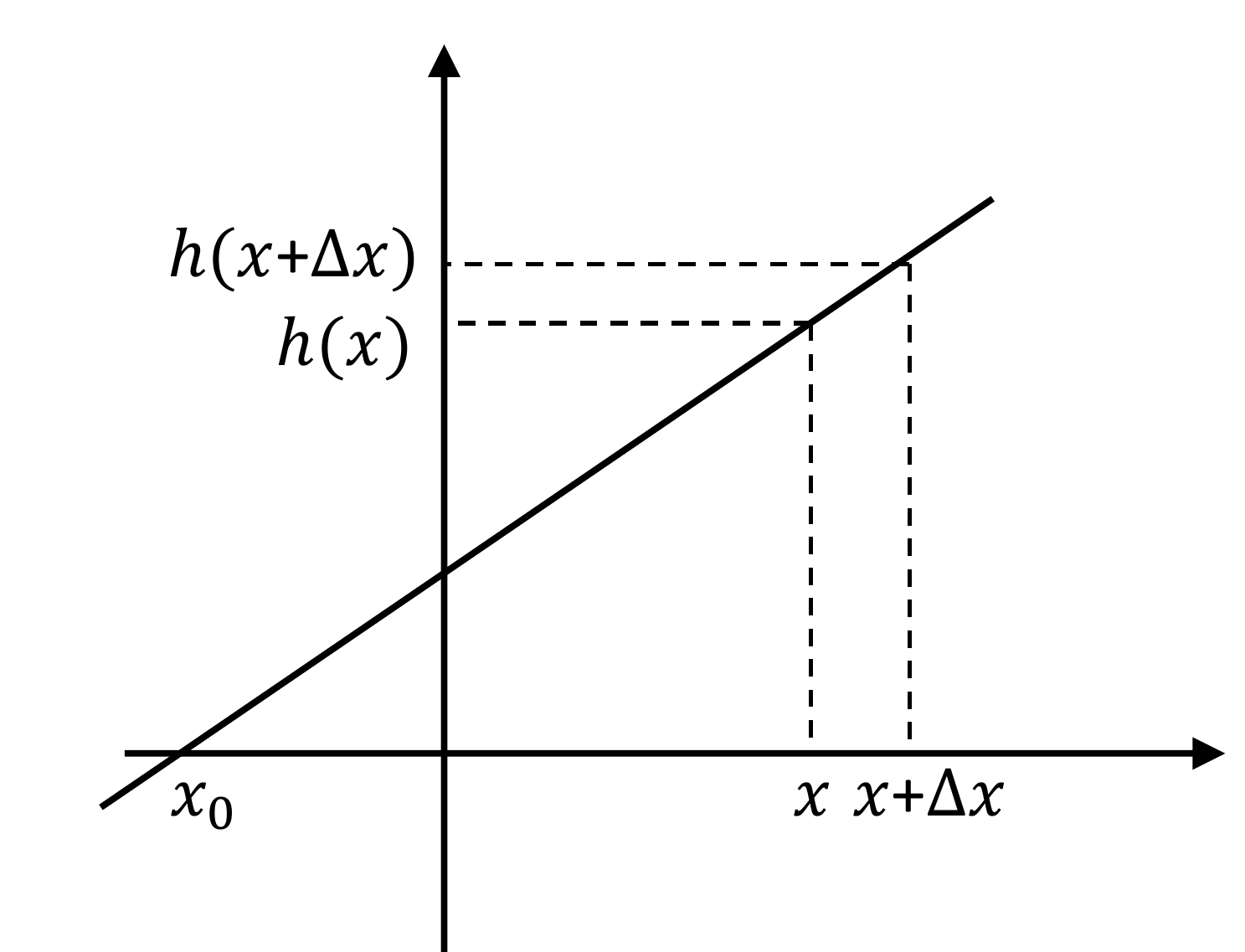}
	\caption{An example of a linear function.}
	\label{fig:proof_1}
\end{figure}

To prove \eqref{equ:proportion}, we take a general linear function $h(\cdot)$ as an example (See Figure~\ref{fig:proof_1}).
For two entries of that function $h(x)$ and $h(x + \Delta x)$, we have:
\begin{equation}
	\centering
	\begin{split}
		\frac{h(x + \Delta x)}{h(x)} & = \frac{x + \Delta x - x_0}{x - x_0} \\
		& = 1 + \frac{\Delta x}{x - x_0} \text{.}\\
	\end{split}
\end{equation}
For a known $x$ and $x_0$, we can get:
\begin{equation}
	\centering
	\frac{h(x + \Delta x)}{h(x)} \approx 1 + H \Delta x \text{,}
\end{equation}
where $H$ is a constant determined by $x_0$ and $x$.
%which proves \eqref{equ:proportion}.

\begin{table*}[]
	\caption{The errors for all the KITTI scenarios that has a time duration longer than $10$s (Unit: degree).}
	\label{tab_all_results}
	\centering
	\footnotesize
	\begin{tabular}{|c|c|c|c|c||c|c|c|c|c|}
		\hline
		Index & Time (s) & Max error & Min error & Mean error & Index & time (s) & Max error & Min error & Mean error\\
		\hline
		0 & 11.0405 & 0.3237 & 0.0021 & 0.1599 & 1 & 15.8007 & 3.2101 & 0.0063 & 1.3192\\
		\hline
		2 & 46.1719 & 2.7596 & 0.0007 & 0.4434 & 3 & 24.0208 & 0.2439 & 0.0011 & 0.1292\\
		\hline
		4 & 14.8103 & 0.3324 & 0.0000 & 0.1022 & 5 & 32.4114 & 3.4554 & 0.1049 & 0.6390\\
		\hline
		6 & 30.6513 & 0.3616 & 0.0018 & 0.1591 & 7 & 11.7006 & 0.0996 & 0.0005 & 0.0659\\
		\hline
		8 & 27.8710 & 3.0607 & 0.0002 & 0.4141 & 9 & 49.7220 & 2.1208 & 0.0001 & 0.3362\\
		\hline
		10 & 82.7736 & 4.0086 & 0.0000 & 0.8431 & 11 & 49.0220 & 2.2211 & 0.0282 & 1.5983\\
		\hline
		12 & 19.3709 & 0.4167 & 0.0023 & 0.1671 & 13 & 44.4319 & 0.9083 & 0.0015 & 0.2834\\
		\hline
		14 & 44.4618 & 3.4304 & 0.0021 & 0.6983 & 15 & 40.3117 & 0.5538 & 0.0178 & 0.2481\\
		\hline
		16 & 13.4506 & 3.0415 & 0.0067 & 0.9338 & 17 & 82.7836 & 2.5787 & 0.0001 & 0.6863\\
		\hline
		18 & 40.6718 & 4.2815 & 0.0614 & 2.5977 & 19 & 12.8108 & 3.7855 & 0.0287 & 1.4215\\
		\hline
		20 & 45.2017 & 0.7892 & 0.0031 & 0.2341 & 21 & 30.2913 & 0.7350 & 0.0022 & 0.4254\\
		\hline
		22 & 37.2817 & 0.2571 & 0.0002 & 0.1038 & 23 & 38.5116 & 1.1222 & 0.0018 & 0.2360\\
		\hline
		24 & 72.6068 & 2.5413 & 0.0004 & 0.3145 & 25 & 58.8447 & 6.1563 & 0.0075 & 2.8520\\
		\hline
		26 & 43.3068 & 4.5927 & 0.0015 & 0.5776 & 27 & 10.2404 & 1.2509 & 0.0001 & 0.4767\\
		\hline
		28 & 39.5816 & 1.6882 & 0.0032 & 0.4421 & 29 & 72.9532 & 2.4156 & 0.0033 & 0.4659\\
		\hline
		30 & 75.3833 & 2.1704 & 0.0012 & 0.5217 & 31 & 35.1015 & 1.5878 & 0.0001 & 0.5401\\
		\hline
		32 & 44.7222 & 6.1210 & 0.0023 & 2.3205 & 33 & 27.6312 & 1.5324 & 0.0000 & 0.5204\\
		\hline
		34 & 49.0221 & 3.7570 & 0.0015 & 0.4451 & 35 & 96.6240 & 2.9939 & 0.0003 & 0.4126\\
		\hline
		36 & 32.1714 & 0.4293 & 0.0006 & 0.1365 & 37 & 23.3907 & 0.3769 & 0.0001 & 0.0680\\
		\hline
		38 & 68.1729 & 4.4276 & 0.0315 & 1.6283 & 39 & 10.8810 & 4.1226 & 0.0018 & 1.3325\\
		\hline
		40 & 38.8837 & 0.4263 & 0.0004 & 0.2748 & 41 & 19.1818 & 0.0583 & 0.0006 & 0.0317\\
		\hline
		42 & 21.5621 & 0.2063 & 0.0029 & 0.1426 & 43 & 11.3010 & 0.6694 & 0.0043 & 0.3790\\
		\hline
		44 & 36.3736 & 0.1897 & 0.0002 & 0.0699 & 45 & 14.9215 & 0.5025 & 0.0006 & 0.2515\\
		\hline
		46 & 35.1455 & 0.1579 & 0.0003 & 0.0449 & 47 & 16.3082 & 0.1612 & 0.0006 & 0.1094\\
		\hline
		48 & 109.8861 & 1.3053 & 0.0006 & 0.2147 & 49 & 28.9393 & 0.1707 & 0.0007 & 0.0493\\
		\hline
		50 & 287.6342 & 5.6801 & 0.0001 & 2.3777 & 51 & 114.4888 & 3.5636 & 0.0018 & 1.7491\\
		\hline
		52 & 114.8479 & 4.7994 & 0.0024 & 1.2012 & 53 & 537.7792 & 5.2659 & 0.0009 & 1.3207\\
		\hline
		54 & 165.3066 & 3.5704 & 0.0011 & 1.3171 & 55 & 126.8774 & 5.1563 & 0.0000 & 0.7062\\
		\hline
		56 & 470.8366 & 11.4645 & 0.0003 & 2.4677 & 57 & 483.4850 & 5.4015 & 0.0001 & 1.0578\\
		\hline
		58 & 121.1863 & 3.2177 & 0.0187 & 2.3184 & 59 & 86.6303 & 1.3729 & 0.0003 & 0.3054\\
		\hline
	\end{tabular}
\end{table*}

\section*{Acknowledgment}

This project/research has received funding from the European
Union’s Horizon 2020 Framework Programme for Research and
Innovation under the Specific Grant Agreement No.945539 (Human Brain Project SGA3).

\ifCLASSOPTIONcaptionsoff
  \newpage
\fi



%\bibliographystyle{IEEEtran}
\bibliography{icra2021}
\bibliographystyle{plain}

\balance
\end{document}